\documentclass{article}

\usepackage{microtype}
\usepackage[moderate]{savetrees}
\usepackage{graphicx}
\usepackage{booktabs} 

\usepackage[hyphens]{url}
\usepackage{hyperref}
\usepackage{amssymb}
\usepackage{amsmath}
\usepackage{cleveref}
\usepackage{xcolor}
\usepackage{subcaption}
\usepackage{bbm}
\usepackage[title]{appendix}
\usepackage{multirow}

\usepackage[acronym]{glossaries}
\glsdisablehyper
\newacronym{ml}{ML}{Machine Learning}
\newacronym{mdp}{MDP}{Markov Decision Process}
\newacronym{pomdp}{POMDP}{Partially Observable Markov Decision Process}
\newacronym{rl}{RL}{Reinforcement Learning}
\newacronym{a2c}{A2C}{Advantage Actor Critic}
\newacronym{a3c}{A3C}{Asynchronous Advantage Actor Critic}
\newacronym{gae}{GAE}{Generalized Advantage Estimation}
\newacronym{dqn}{DQN}{Deep Q Network}
\newacronym{ddqn}{DDQN}{Double Deep Q Network}
\newacronym{dnn}{DNN}{Deep Neural Network}
\newacronym{abr}{ABR}{Adaptive Bit Rate}
\newacronym{bba}{BBA}{Buffer Based Approach}
\newacronym{qoe}{QoE}{Quality of Experience}
\newacronym{cf}{CF}{Catastrophic forgetting}

\usepackage[accepted]{mlsys2023}

\usepackage{enumitem}

\mlsystitlerunning{Demystifying Reinforcement Learning in Time-Varying Systems}

\begin{document}

\twocolumn[
\mlsystitle{Demystifying Reinforcement Learning in Time-Varying Systems}

\mlsyssetsymbol{equal}{*}

\begin{mlsysauthorlist}
\mlsysauthor{Pouya Hamadanian}{mit}
\mlsysauthor{Malte Schwarzkopf}{brown}
\mlsysauthor{Siddartha Sen}{microsoft}
\mlsysauthor{Mohammad Alizadeh}{mit}
\end{mlsysauthorlist}

\mlsysaffiliation{mit}{Computer Science and Intelligence Lab, Massachusetts Institute of Technology, Cambridge, Massachusetts, USA}
\mlsysaffiliation{brown}{Computer Science Department, Brown University, Providence, Rhode Island, USA}
\mlsysaffiliation{microsoft}{Microsoft Research, New York City, New York, USA}

\mlsyscorrespondingauthor{Pouya Hamadanian}{pouyah@mit.edu}

\mlsyskeywords{Reinforcement Learning, Non-stationary, Non-stationarity, System Optimization, Load Balancing, Adaptive Bit Rate, Machine Learning for Systems, Multiple experts, Safety, Context detection, MLSys}

\vskip 0.3in

\begin{abstract}

Recent research has turned to Reinforcement Learning (RL) to solve challenging decision problems, as an alternative to hand-tuned heuristics. RL can learn good policies without the need for modeling the environment's dynamics. Despite this promise, RL remains an impractical solution for many real-world systems problems. A particularly challenging case occurs when the environment changes over time, i.e. it exhibits {\em non-stationarity}. In this work, we characterize the challenges introduced by non-stationarity, shed light on the range of approaches to them and develop a robust framework for addressing them to train RL agents in live systems. Such agents must explore and learn new environments, without hurting the system's performance, and remember them over time. To this end, our framework (i) identifies different environments encountered by the live system, (ii) triggers exploration when necessary, (iii) takes precautions to retain knowledge from prior environments, and (iv) employs safeguards to protect the system's performance when the RL agent makes mistakes. We apply our framework to two systems problems, straggler mitigation and adaptive video streaming, and evaluate it against a variety of alternative approaches using real-world and synthetic data. We show that all components of the framework are necessary to cope with non-stationarity and provide guidance on alternative design choices for each component.

\end{abstract}
]

\printAffiliationsAndNotice{}  

\section{Introduction}
\label{sec:intro}
\noindent
{\em --- The only constant is change. (Heraclitus 500 B.C.)}

Deep reinforcement learning has been proposed as a powerful solution to complex decision making problems~\cite{mnih2013playing, alphazero}. In systems, it has recently been applied to a wide variety of tasks, such as adaptive video streaming~\cite{mao2017neural}, congestion control~\cite{jay2019internet}, query optimization~\cite{krishnan2019learning,marcus2019neo}, scheduling~\cite{mao2019learning}, resource management~\cite{mao2016resource}, device placement~\cite{pmlr-v80-gao18a}, and others. Reinforcement learning is particularly well-suited to these problems due to the abundance of data and its ability to automatically learn a good policy in the face of complex dynamics and objectives.

Fundamentally, reinforcement learning trains an agent by giving it feedback for decisions it makes while interacting with an environment. This interaction can occur in a controlled environment, such as a simulation or testbed, or in a real environment, such as a live deployment. While using a controlled environment seems like an attractive choice---e.g., it is data-efficient and less invasive---policies trained in this manner do not perform well in the real world~\cite{yan2020learning} due to differences between the simulation and deployment environments. To circumvent this problem, therefore, it is appealing to carry out reinforcement learning \emph{in-situ}, i.e., by interacting with the live system. However, training an agent on a live system introduces several challenges.

First, an RL agent interacting with a system for the first time will inevitably make many mistakes as it explores possible strategies. This could endanger the system and lead to significant performance loss. Second, real-world systems are time-varying and exhibit considerable non-stationarity: for example, shifting workload patterns, network characteristics, hardware characteristics, and so on. Adapting to new conditions requires {\em continual} reinforcement learning, which is non-trivial~\cite{khetarpal2020towards}. Simply retraining the agent on new observed data is not sufficient. To learn new behaviors, the agent needs to re-explore the environment when it changes, potentially degrading system performance. Furthermore, training in a new environment may lead to forgetting older ones, a problem known as {\em catastrophic forgetting}~\cite{MCCLOSKEY1989109}. Left unattended, this would necessitate retraining the agent on every change in the system, even on environments that have been observed and trained on numerous times before.

An ideal reinforcement learning agent would explore exactly as much as necessary, and exploit its knowledge from that point onwards, until the environment shifts to a new dynamic. This agent would retain all of its knowledge forever, and would never need to retrain on the same environment twice. Moreover, it would explore in a safe manner, limiting system impact. 
Unfortunately, no off-the-shelf training method satisfies all these goals. While the literature provides point solutions to different aspects of the problem (e.g., safety, catastrophic forgetting, exploration, etc.), there is no holistic approach to designing practical systems in the real-world. Most current techniques are evaluated in simple benchmark tasks that fail to capture the nuances and full breadth of challenges of a real deployment. For example, our results show that state-of-the-art {\em change-point} detection techniques used to combat catastrophic forgetting fail under real-world non-stationarity. 

Our goal is to conduct a comprehensive analysis of the challenges of online reinforcement learning in time-varying computer systems and shed light on the design-space for mitigating these challenges (\S\ref{sec:non_stat_approach}). To explore this design space, we propose a framework to realize an ideal \gls{rl} agent described above (\S\ref{sec:framework_challenges}). This framework includes the key elements that we believe are necessary in practice: (i) an environment detector that detects changes in operating conditions, (ii) a method to trigger exploration when necessary, (iii) a method to retain all prior knowledge, and (iv) a safety monitor to check for unsafe conditions and revert the system to known safe states. As mentioned earlier, some techniques have appeared in prior work (\S\ref{sec:related_works}) for each of the elements in this framework, but our goal is to understand how these techniques work together in practice, and provide guidance on how designers should reason about alternative design choices.

We conduct case studies on two systems problems, using a combination of real-world and synthetic data: straggler mitigation in job scheduling (\S\ref{sec:load_balance}), and bitrate adaptation in video streaming (\S\ref{sec:abr}). Our key findings are: (i) Environment detection is a crucial ingredient to avoid catastrophic forgetting in online RL, but existing techniques that rely on generic methods (e.g., model-based likelihood methods, etc.)~\cite{alegre2021minimum} to detect changes perform poorly in real systems. Incorporating domain knowledge about key features and system behavior can significantly improve performance. (ii) Off-policy approaches that try to sidestep environment detection by training on all historical data (e.g., Experience Replay~\cite{mnih2013playing}) often perform poorly compared to on-policy learning methods that train an ensemble of expert policies tailored to specific environments. Besides the known limitations of off-policy learning~\cite{haarnoja2018soft}, we show that their performance is very sensitive to strategies used to retain historical data. (iii) Simple safeguards that switch to default policies when the system deviates significantly from expected behavior are quite effective but should be designed based on domain knowledge about system dynamics. 

Our study shows that our framework, when instantiated carefully, can avoid notable failures and approximate the ideal RL agent in real-world systems. We hope that our results can serve to demystify how to apply reinforcement learning to time-varying systems in the wild.

\section{Preliminaries}
\label{sec:background}

We first discuss how to formulate decision making problems in an \gls{rl} setting, and then briefly discuss two types of \gls{rl} algorithms we use throughout the paper.

\subsection{\gls{mdp}}
\label{subsec:mdp}

An \gls{rl} problem consists of an \emph{environment}, which is a dynamic control system modeled as an \gls{mdp}~\cite{sutton_intro_rl}, and an \emph{agent}, which is the entity affecting the environment through a sequence of decisions. The agent observes the environment's \emph{state}, and decides on an \emph{action} that is suitable to take in that state. The environment responds to the action with a \emph{reward} and then transitions to a new state. 
Formally, at time step $t$ the environment has state $s_t \in \mathcal{S}$, where $\mathcal{S}$ is the space of possible states. The agent takes action $a_t\in \mathcal{A}$ from the possible space of actions $\mathcal{A}$, and receives feedback in the form of a scalar reward $r_t(s_t, a_t): \mathcal{S}\times\mathcal{A} \rightarrow \mathbb{R}$. The environment's state and the agent's action determine the next state, $s_{t+1}$, according to a transition kernel, $T(s_{t+1}|s_t, a_t)$, which provides the distribution of the next state conditioned on the current state and action. Finally, $d_0$ defines the distribution over initial states ($s_0$). An \gls{mdp} is fully defined by the tuple $\mathcal{M}=(\mathcal{S}, \mathcal{A}, T, d_0, r)$.\footnote{In many problems, the agent cannot observes the full state $s_t$ of the environment. These problems can be modeled as \gls{pomdp}s, in which the agent observes a limited observation $o_t$ instead.} The goal of the agent is to optimize the \emph{return}, i.e. a discounted sum of rewards $R_0 = \sum_{t=0}^{\infty} \gamma^t r_t$.

\subsection{\gls{rl} algorithms}
\label{subsec:rl_algo}

In this paper we focus on two classes of model-free \gls{rl} algorithms~\cite{sutton_intro_rl}: \emph{on-policy} and \emph{off-policy} algorithms. Model-free algorithms are the most common approach in \gls{rl}-based systems, since they avoid the need to model the environment's dynamics, which is difficult (and often infeasible) in real-world systems~\cite{mao2016resource}.

\textbf{On-policy \gls{rl}}:
To train an \gls{rl} agent with an on-policy method, we start by deploying an initial policy to interact with the system. Each interaction creates a sample {\em experience} (a tuple comprising of a state, action, reward, and next state), and once enough samples are collected, the agent is trained for a single step. Importantly, the collected samples are discarded once they have been used to update the policy. This is the main limitation of on-policy methods: a policy can only be trained on samples created by the same policy. Operationally, the implication is that such methods require a substantial amount of interaction with the live system for training, and they are inherently prone to forgetting past behaviors because they discard old experience data.
In this paper, we consider \gls{a2c}, a prominent on-policy algorithm~\cite{mnih2016asynchronous} based on policy gradients.
We refer the reader to \S\ref{app:a2c} in the appendix for a brief explanation of this approach.

\begin{figure*}[t!]
    \includegraphics[width=\linewidth]{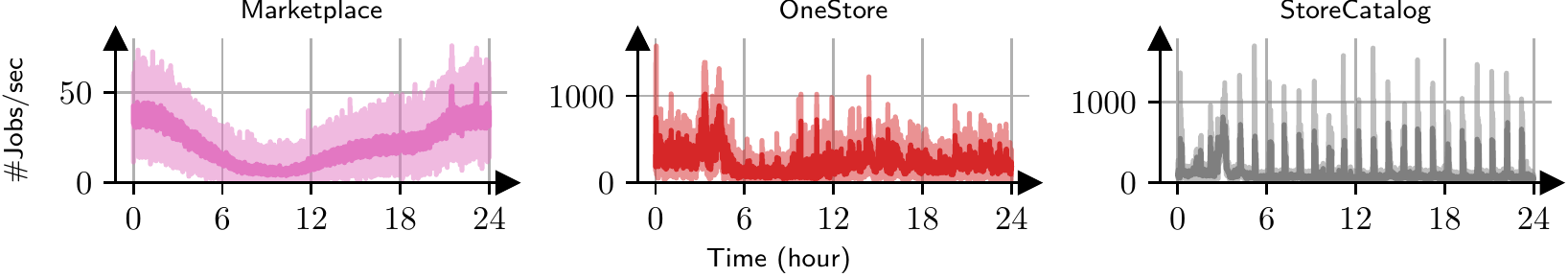}
    \vspace{-19pt}
    \caption{Visualizing non-stationarity in three example traces from a production web framework cluster. Shaded regions show actual arrival rate in 500 ms intervals, and solid lines are exponentially weighted averages ($\alpha=0.9$).}
    \label{fig:ex_workloads}
    \vspace{-12pt}
\end{figure*}

\textbf{Off-policy \gls{rl}}:
Unlike on-policy methods, off-policy approaches can train with samples from a policy different from the one being trained, such as historical data (collected with an earlier policy used during training). Off-policy methods maintain a record of previous interactions (also called an experience replay buffer) and use them for training. Naturally, these methods are more sample-efficient than their on-policy counterparts, but they are also known to be more prone to unstable training, suboptimal results, and hyper-parameter sensitivity~\cite{haarnoja2018soft,duan2016benchmarking, gu2016q}. 
The most popular off-policy deep RL method is the \gls{dqn} algorithm~\cite{mnih2013playing}; we consider a variant of it in this paper. For more explanations regarding \gls{dqn}, refer to \S\ref{app:dqn} in the appendix.
\section{RL in non-stationary systems}
\label{sec:non_stat_approach}

We consider non-stationary systems that are subject to time-varying operating conditions, e.g.,  request arrival rate in a micro-service, network capacity in a video streaming application, cross-traffic on a network path, etc. We will refer to the non-stationarity aspect of the system as the {\em workload}, signifying that it is often an exogenous factor impacting the dynamics of the system. From a modeling standpoint, one can think of the current active workload as affecting the \gls{mdp} dynamics. Actions in the same states but under different workloads have different consequences, and the optimal sequence of actions differs for each workload. Online \gls{rl} algorithms are challenged in such environments; when they train, they adapt to the specific workload they experienced and perform poorly in a new one. Unfortunately, the solution is not as simple as retraining on new data. There are three challenges that we have to face.

\begin{figure*}[t!]
    \begin{subfigure}{0.34\textwidth}
        \includegraphics[trim=0.2cm 0.1cm 15cm 0.1cm, clip, width=\linewidth]{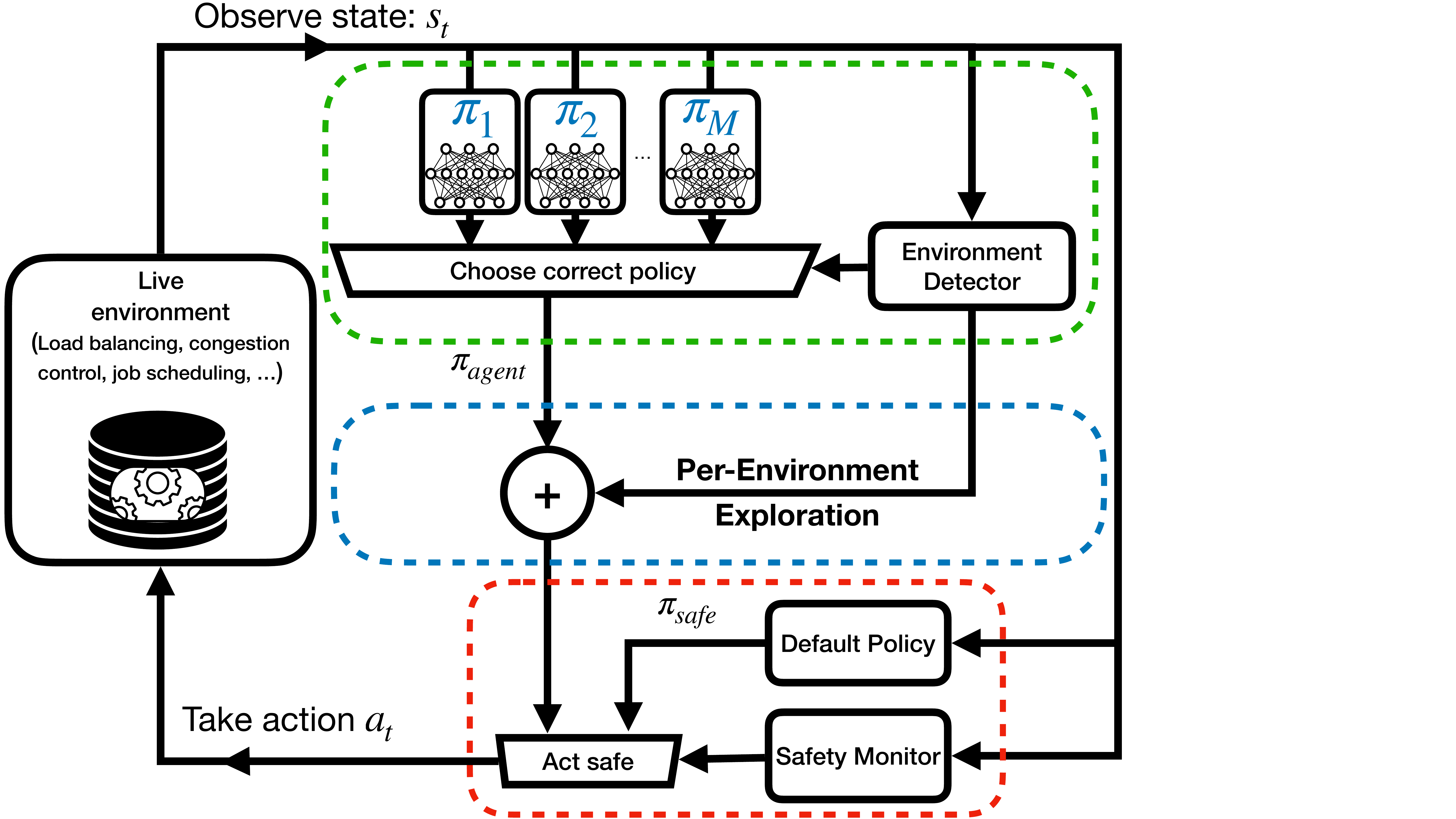}
        \caption{}
        \label{fig:frame_view}
        \vspace{-12pt}
    \end{subfigure}
    \hspace{7mm}
    \begin{subfigure}{0.63\textwidth}
        \hspace*{-0.15cm}
        \includegraphics[trim=0.35cm 0.4cm 0 0.4cm, clip, width=\linewidth]{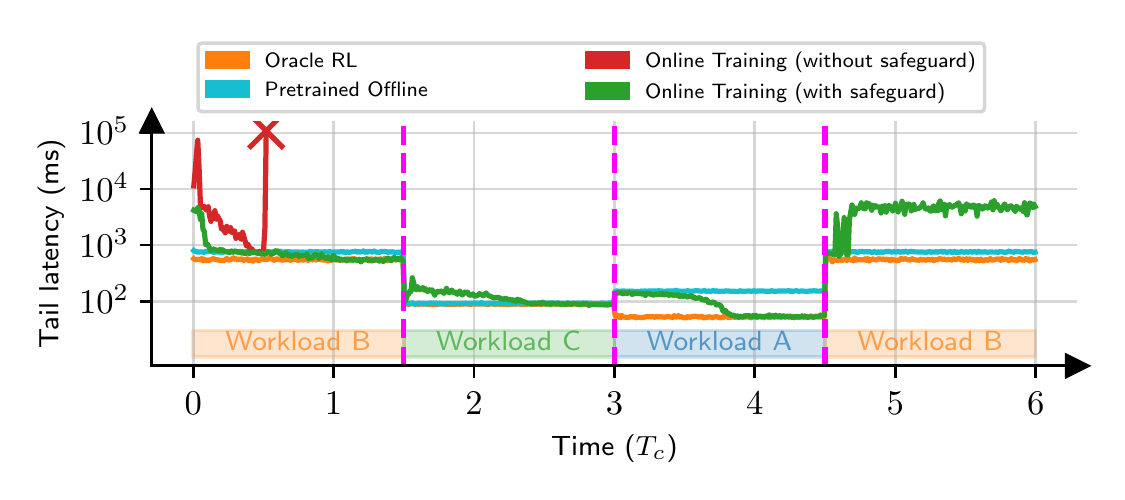}
        \caption{}
        \label{fig:challenges_time_series}
        \vspace{-12pt}
    \end{subfigure}
    \caption{\textbf{(a)} Framework overview for non-stationary \gls{rl}. {\color{green!50!black} Green} components mitigate catastrophic forgetting, {\color{blue} Blue} components concern per-environment exploration, and {\color{red} Red} components deal with safety. \textbf{(b)} An example demonstrating the challenges of learning a straggler mitigation policy in an environment with time-varying workloads. Curves denote the $95^\text{th}$ percentile job latency over 5 minutes windows. $T_C$ denotes the convergence time when training on a stationary workload. Active workloads are marked and colored below the curves and vertical dashed lines denote workload changes.
    We include a pretrained agent, which performs poorly on workload A that is far from its training workloads.
    }
    \vspace{-6pt}
\end{figure*}

The main obstacle is not forgetting the past. Even if we correctly retrain for new workloads, the adaptation to previous workloads is lost, otherwise known as \gls{cf}.
Specifically, when a neural network is updated based on experiences derived from one workload for a long time, it tends to overfit to the data distribution of that workload, and forgets behaviors learned from earlier data~\cite{MCCLOSKEY1989109, PARISI201954, Atkinson_2021}. Despite research on this topic, there is no general approach to avoiding \gls{cf} in non-stationary \gls{rl}~\cite{khetarpal2020towards}.

Non-stationary \gls{rl} research mainly focuses on the task-based setting, where each workload is assigned an index that is known at interaction time. In practice, no such index is available, but we can derive indices ourselves with generic change-point detection methods~\cite{Padakandla_2020, alegre2021minimum, rlcd, doya2002multiple, ren2022hdpcmdp}. Such methods learn a model of \gls{mdp} transition dynamics and whenever the learnt model doesn't align with currently observed transitions, they declare a new task. However, real systems environments are not as well-behaved as these methods require; there is great variance in transitions, and workloads are quite unpredictable in how they change. \Cref{fig:ex_workloads} shows the request arrival rate to three services from a production cluster operated by a large web service provider. Note how the workload can change smoothly, abruptly, or with no discernible pattern. While generic approaches are not robust for such workloads (as we demonstrate in \S\ref{sec:real_results}), detecting task-indices is a reasonable approach, and we demonstrate how incorporating domain knowledge can improve robustness. 

An entirely different method is using experience replay. The idea is that off-policy approaches can train on stale data from all past workloads. These approaches are commonly used in applied \gls{rl}, e.g. in 5G research~\cite{sritharan2020study, gu2021knowledge, sadeghi2017optimal, 9839177}, but are poor-performing, as we show extensively in \S\ref{subsec:lb_exps}, \S\ref{sec:real_results} and \S\ref{app:lb_off_scaling}. There are three main reasons: (1) To solve \gls{cf}, off-policy methods still need a buffering strategy, but none are robust in practice. (2) All \gls{rl} algorithms need to explore when facing new workloads, which means they still need task indices to know when to re-explore. (3) It is a well-known fact that when sample-complexity is not a concern, off-policy methods do not perform as well as on-policy approaches~\cite{haarnoja2018soft,duan2016benchmarking, gu2016q}.

Besides \gls{cf}, there are two other issues. First, \gls{rl} training requires an exploration phase, where the agent tries good and bad actions alike. During this phase the agent performs poorly, and is only useful after it ends. Thus, to train on each new workload we need to re-initiate exploration. Of course, when available, we can use inferred task-indices to control exploration. But when no such task-index exists, such as in off-policy approaches, exploration is non-trivial. An alternative approach is maintaining a small level of exploration all the time. However, prior work~\cite{ahmed2019understanding} and our experiments showed that this leads to poor policies and is difficult to tune appropriately. 

Finally, as \gls{rl} methods explore, they can make bad decisions and violate performance requirements. This is obviously problematic a live system and we need a way to restrict exploration to safe levels during the training period.

\section{Framework Overview}
\label{sec:framework_challenges}

Successful deployment of \gls{rl} in non-stationary systems is challenging, but possible with the correct structure. In this section, we outline a framework, visualized in \Cref{fig:frame_view}, that serves as a blueprint for training RL agents in a live system. We illustrate the importance of each component via a motivating example of mitigating stragglers in an online service by ``hedging" requests---i.e., replicating a request when a response doesn't arrive within a timeout, covered in depth in \S\ref{subsec:lb_setup}. This system is subject to time-varying workloads that determine how the agent's decisions affect the system: e.g., a certain timeout threshold for request hedging may be beneficial for one workload, but lead to congestion in another. 
For simplicity, we switch between several workloads here, and evaluate realistic scenarios in \S\ref{sec:real_results}. The active workload in each period of time is indicated at the bottom of \Cref{fig:challenges_time_series}. The curves in the figure denote the tail of job latencies over time (note the logarithmic scale), with the objective of minimizing this latency.

Our framework consists of three key modules: a safety monitor, a default policy, and an environment detector, all defined by the system designer. This framework is a template; it guides the system designer to incorporate domain knowledge where it matters the most. The \emph{safety monitor} checks for violation of a safety condition as the agent interacts with the environment. 
The \emph{default policy} is an existing scheme that, when activated in an unsafe regime, can return the system to safe operation. 
The \emph{environment detector} aims to detect which environment is active at any given time. It typically takes in a small set of features (selected by the system designer) and uses them to detect changes in the environment. For instance, to detect changes in the workload, we might use features such as job arrival rates and job types.

{\color{red}\textbf{Safety}}:
As discussed earlier, online exploration in a live system can degrade performance, and possibly drive the system to disastrous states. Take the ``online training without safeguards" curve in \Cref{fig:challenges_time_series} as an example. Without a safety mechanism to keep things in check, the tail latency can shoot up and even grow without bound. Inspired by recent work~\cite{mao2019towards}, we approach safety in our framework using safeguards that limit certain actions when the environment is unsafe. For instance, we can disable hedging when tail latency exceeds a certain threshold in our example (see \S\ref{subsec:lb_setup} for details). Generally, as shown in \Cref{fig:frame_view}, a safety monitor controls whether the agent or a default policy decides the next action. When in an unsafe state, the default policy is activated, driving the system to a safe region before control is given back to the \gls{rl} agent to resume training. This limits the performance impact of exploration or other misbehaviors by the \gls{rl} agent (e.g., transient policy problems during training), as seen with ``online training with safeguards" in \Cref{fig:challenges_time_series}: the tail latency remains bounded throughout.

{\color{blue}\textbf{Per-Environment Exploration}}: To learn new behaviors when the environment (workload) changes, we need to explore again. However, perpetually exploring on every change is not desirable, because even with safeguards, exploration incurs a performance cost. Ideally, we should explore sparingly, i.e., once for each new workload. Thus, for each observed environment, we initiate a one-time exploration. The choice to explore is driven by the environment detection signal, as shown in \Cref{fig:frame_view}.
In \Cref{fig:challenges_time_series}, ``online training with safeguards" tracks environment changes and initiates an exploration phase that lasts for one $T_c$ (period of convergence, see \Cref{fig:challenges_time_series}) the first time it encounters a new workload.
Accordingly, the tail latency initially degrades when a new workload begins, and then improves as the agent shifts from exploration to exploitation.

{\color{green!50!black}\textbf{Catastrophic forgetting}}: \Cref{fig:challenges_time_series} shows another challenge for online \gls{rl}. Ideally, when workload B appears a second time, the agent should be able to immediately exploit its past knowledge. But the results show that ``online training with safeguards" fails to remember what it had learnt. This is due to \gls{cf}.
We adopt a simple solution that avoids this problem altogether by training different policies for each environment, also referred to as multiple ``experts''~\cite{rusu2016policy}. We select the appropriate expert to use and train in each environment using the environment detector. Our results suggest that this approach can be quite effective, even when the environment detector isn't perfect and makes mistakes.

Multiple experts are not the only solution for \gls{cf} and a rich variety of methods have been proposed~\cite{atkinson2021pseudo, rusu2016progressive, kirkpatrick2017overcoming}. Such methods might enable sharing knowledge learned from one workload for training another. However, multiple experts has the advantage of being robust and simple to design and interpret.
\section{Case Study: Straggler Mitigation}
\label{sec:load_balance}

\label{subsec:lb_setup}

\begin{figure}[t!]
  \includegraphics[trim=11cm 0.25cm 1.5cm 2.25cm, page=1, clip, width=0.8\linewidth]{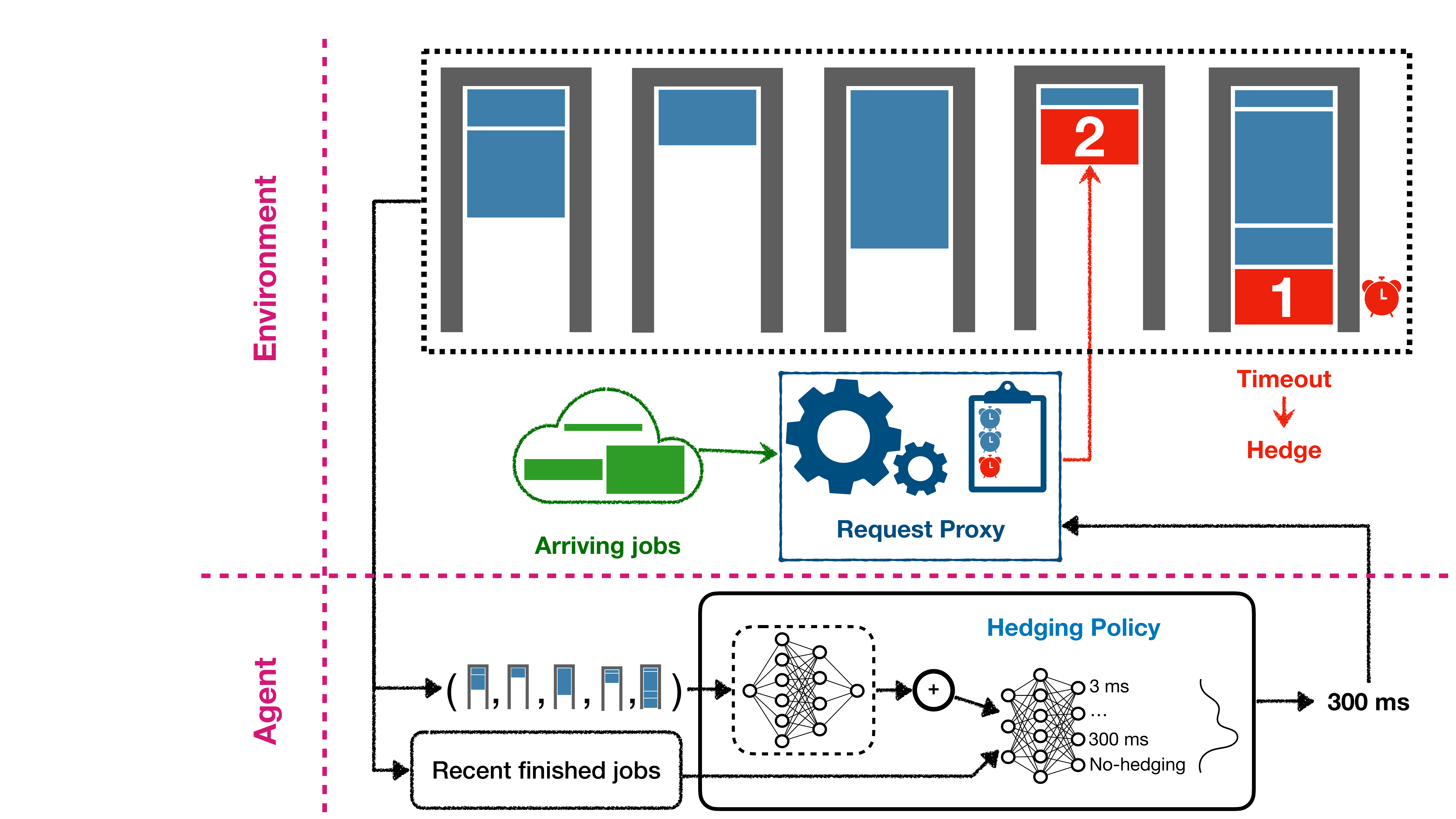}
  \caption{Illustration of a request proxy with hedging.}
  \label{fig:lbalance_fig}
  \vspace{-20pt}
\end{figure}

We consider a simulated request proxy with hedging (\Cref{fig:lbalance_fig}). In this environment, a proxy receives and forwards requests to one of $n$ servers. The servers process requests in their queues one by one. To load balance, the proxy sends the request to the server with the shortest queue.
To respond to a request, the server launches a job that requires a nominal processing time, which we will refer to as its size. The size of a job is not known prior to it being processed. Henceforth we will use the terms ``request'' and ``job'' interchangeably. 

In a real system, some jobs may incur a slow down and take longer than the nominal time, e.g., due to unmet dependencies, IO failure, periodic events such as garbage collection~\cite{tailscale}, noisy neighbours~\cite{Pu2010UnderstandingPI}, etc. These slowdowns also affect other jobs further in the queue. The effect is especially pronounced at the tail of job latency, which is the time between the job's arrival and its completion.  To simulate such behaviors in our environment, we inflate the processing time of a job relative to the nominal value by a factor of $k$ with a small probability $p$. We use $k=10$ and $p=0.1$ in our experiments.

To mitigate the effect of slowdowns, if a job has not completed by a timeout, the proxy ``hedges'' it by sending a duplicate request to another server. This duplication only happens once per job. The job completes when either request (the original or its duplicate) finishes. If the hedged request finishes faster than the original one, the job's latency is reduced. Our goal is to minimize the $95^\text{th}$ percentile latency, and our control decision is the hedging timeout. The hedging policy gets to pick among a set of six timeout values, ranging from $3^{ms}$ to $300^{ms}$, or alternatively to do no hedging. 

There is an inherent trade off in selecting a hedging timeout. A low timeout leads to more jobs being hedged, possibly reducing their latency, but it also creates more load on the system, resulting in bigger queues. An \gls{rl} agent can learn the optimal threshold, which depends on the amount of congestion (e.g., queue sizes) in the system and the workload (e.g., the job arrival rate and job sizes). Since workloads change over time, this environment is non-stationary.

As nothing is known about jobs prior to processing, it doesn't help to make individual decisions on a per-job basis. Instead, the agent chooses one hedging timeout for all jobs that arrive within a time window (500~ms in our experiments). It makes these decisions using the following observations: \textbf{(a)} instantaneous server queue sizes, \textbf{(b)} average server queue sizes over the last time window, \textbf{(c)} average and max of job processing times and rate of incoming jobs within $m=4$ time windows, \textbf{(d)} the average load in the last time window, and \textbf{(e)} whether a safeguard is active (more details below). The reward is the negated 95\textsuperscript{th} percentile latency of jobs in that window. 

With $n$ servers, the observation includes $2n$ dimensions regarding instantaneous and average queue statistics. These statistics are in essence a set, and the order they appear in the observation does not matter, e.g., if queue 4 comes before queue 8 or vice versa. We therefore use the DeepSets~\cite{zaheer2018deep} architecture to create permutation-invariant neural network models that exploit this structure. At a high level, this architecture works by learning an embedding function applied to each queue's statistics, and summing the resulting embeddings to represent the set of queue statistics. 
\S\ref{app:lb_train} has a full description of our training setup and environment.

\textbf{Safeguard}:
We use a safeguard policy~\cite{mao2019towards} to improve transient performance and stabilize training. The safeguard overrides the agent when at least one queue builds up past a threshold (50) and disables hedging while in control. It relinquishes control back to the \gls{rl} agent once all queue sizes are below a safe threshold (3). 

\textbf{Workloads}:
We use traces from a production web framework cluster at AnonCo, collected from a single day in February 2018. The framework services high-level web requests to different websites and storefront properties and routes them to various backend services (e.g., product catalogs, billing, etc.). The traces are noisy, heavily temporally-correlated, and change considerably over time, as \Cref{fig:app_offline_visual,fig:app_workload_time} in the appendix demonstrate. For more details, see \S\ref{app:workload}.

\textbf{Evaluation metric}:
For evaluation, we calculate the $95^\text{th}$ percentile latency in 5 minute windows. We use this tail latency metric in all experiments, by either plotting it as a timeseries, or by visualizing its distribution with boxplots.

\subsection{Experiments with controlled non-stationarity}
\label{subsec:lb_exps}

\begin{figure}[t]
    \includegraphics[trim=0 0.3cm 0 0.3cm, clip, width=\linewidth]{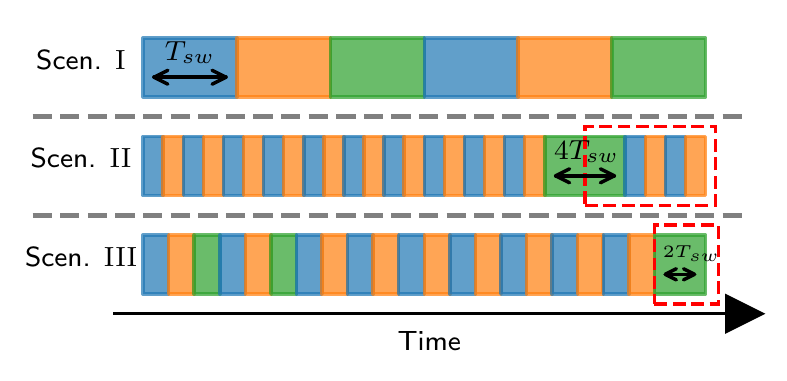}
    \caption{We evaluate three non-stationary scenarios. Scenario I: the system cycles through three different workloads, each active for $T_{sw}$ at a time. Scenario II: two workloads periodically switch until a new workload shows up. Scenario III: three workloads periodically switch, but one becomes inactive for a long time, then reoccurs.}
    \label{fig:scenario_all}
    \vspace{-20pt}
\end{figure}

First, to better understand how non-stationarity affects online RL strategies, we use a simple controlled setup to inflict different levels of non-stationarity. We consider a scenario where several workloads (\emph{Workload A}, \emph{B} and \emph{C} from \Cref{fig:app_workload_time}) change according to a schedule (scenario I in \Cref{fig:scenario_all}). Specifically, workloads change in a periodic and cyclic manner, each active for a period of $T_{sw}$ at a time. In all experiments, we run this scenario with different permutations of workloads (e.g., \emph{ABC}, \emph{BCA}, \emph{CBA}, etc.), as the order of workloads can affect RL training schemes. The period between switches, $T_{sw}$ is relative to convergence time $T_c$ of an RL agent trained for only one workload.

\noindent \textbf{Exploration is critical with new workloads:}
In \S\ref{sec:framework_challenges} we stated that every time the workload changes, we need to explore again. 
This requires designing an environment detector and switching to the exploration phase whenever a change is observed. 
In this environment, detecting changes is relatively simple. Employing our domain knowledge, we recognize two key features: the job arrival rate and the job sizes to characterize the workload (as shown in \Cref{fig:app_workload_time}). By tracking these features, we can cluster our workloads and train a classifier for them. Of course, the classifier may not be perfect, e.g., the clusters may overlap and not be completely separable. But our experiments show that this simple approach is adequate. It is possible to design more complex features and clustering schemes, but the exact approach to designing an environment detector is not our focus in this work.

\begin{figure}[t]
    \includegraphics[trim=0 0.3cm 0 0.1cm, clip, width=\linewidth]{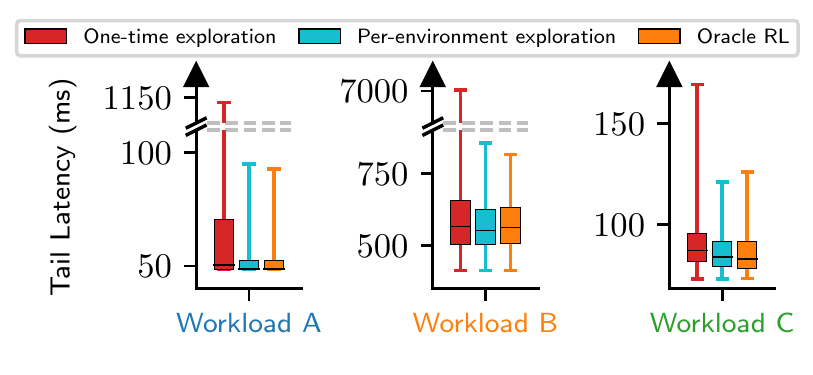}
    \caption{Tail latency distribution after convergence; Each box represents one of the workloads. Whiskers show 99 and 1 percentiles. Upper, middle and lower edge of boxes show 75, 50, and 25 percentiles. Each workload is given enough time to converge ($T_{sw} = 2.25 T_c$).}
    \label{fig:on_policy_reset_box}
    \vspace{-18pt}
\end{figure}

\Cref{fig:on_policy_reset_box} demonstrates the effect of exploring once for each workload. Each boxplot shows tail latency distributions for one of three workloads (across different experiments in which we permute the workload order). The boxes show the $25^\text{th}$, median and $75^\text{th}$ percentiles of the distributions and whiskers denote $1^\text{st}$ and $99^\text{th}$. While one-time exploration (for $T_c$ time) fails to handle new workloads, per-environment exploration converges to a good policy for each workload. \Cref{fig:app_on_policy_reset_timeseries} in the appendix provides the time-series of tail latency with these approaches.

\noindent \textbf{Catastrophic Forgetting:}
The largest obstacle to online learning is \gls{cf}. As observed in \Cref{fig:challenges_time_series}, sequential training will cause a single model to forget its previous training. 
We evaluate two techniques to mitigate CF: (1) Providing features as part of the agent's observation that enable it to distinguish between different workloads. By adding these features, the agent can potentially use a single model while learning different behaviors for different workloads.  For this scheme, we use the same features used by our workload classifier. (2) Employing multiple experts, each trained and used for a unique workload, as explained in \S\ref{sec:framework_challenges}. Ideally the expert for workload A will never be used for and trained in workload B, thus never forgetting the policy it learned for workload A. In practice, the environment detector will however make mistakes, but we found that even a $10\%$ error rate will not degrade performance.

\begin{figure}[t]
    \begin{subfigure}{\linewidth}
        \includegraphics[width=\linewidth]{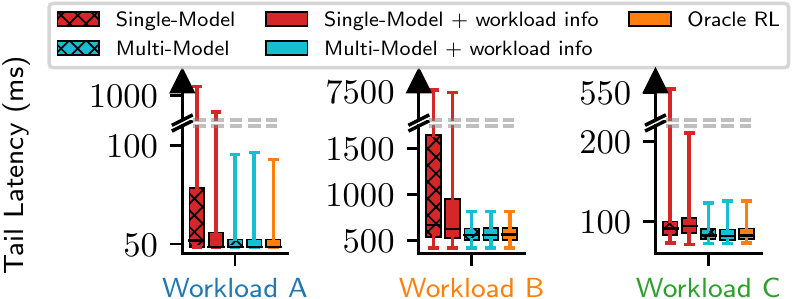}
        \caption{}
        \label{fig:on_policy_CF_slow}
    \end{subfigure}
    \begin{subfigure}{\linewidth}
        \includegraphics[width=\linewidth]{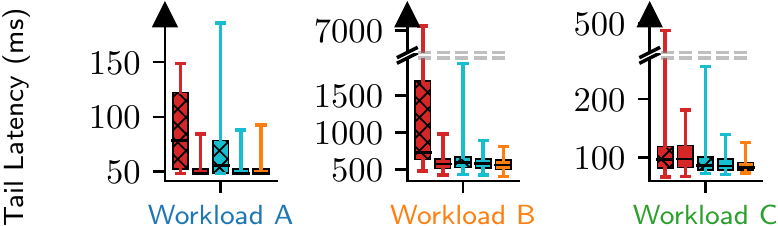}
        \caption{}
        \label{fig:on_policy_CF_fast}
    \end{subfigure}
    \vspace{-20pt}
    \caption{Tail latency distribution after convergence, when either a single model is used or multiple, and when workload statistics are observed or not. \textbf{(a)} $T_{sw}=T_c$ (Slow workload switching), \textbf{(b)} $T_{sw}=0.001~T_c$ (Fast workload switching).}
    \label{fig:on_policy_CF}
    \vspace{-20pt}
\end{figure}

In \Cref{fig:on_policy_CF}, we evaluate multiple experts vs. a single one, and the impact of providing workload features in the observation for both approaches. We consider two different switching periods between workloads: $T_{sw}=T_c$ (slow switching) and $T_{sw}=0.001~T_c$ (fast switching). As a baseline, we also include results for Oracle RL that uses trained policies specific to each workload. 

\Cref{fig:on_policy_CF_slow} concerns situations where workload changes occur at long periods. In these cases multiple experts significantly outperform a single expert, even when workload information is included in the observations. Workload features don't help in this case since their span is limited throughout the convergence interval of the RL algorithm. On the other hand in \Cref{fig:on_policy_CF_fast} where workloads change rapidly, the single model with workload features matches multiple experts. When workloads change at a fast pace, the agent gets to observe samples from all environments (with different workload features) as the RL algorithm converges. Note that in this case the multiple-expert scheme's  performance is worse without workload information; this is due to environment classification errors caused by stale information. However, multiple experts with workload information manages to perform well despite these errors since each expert learns to handle observations from non-matching workloads as well. Overall, multiple experts with workload information is robust in all cases.


\noindent \textbf{Can we avoid catastrophic forgetting with a single model with off-policy methods?} As explained in \S\ref{sec:background}, off-policy methods such as \gls{dqn} can train on historical samples using an experience replay buffer.
Therefore perhaps an off-policy approach can avoid \gls{cf} even with a single model. Using a single model simplifies the agent and could accelerate learning by enabling shared learning across workloads.\footnote{Note that we still need a workload detection scheme for per-environment exploration.}

While appealing, off-policy schemes have their own challenges. In particular, their performance is sensitive to the samples saved in the experience buffer, which must be selected carefully to match the mixture of workloads experienced by the agent over time. Further, as we discuss in \S\ref{app:lb_off_scaling}, using a single model (the Q-network in DQN) across different workloads requires reward scaling to ensure some workloads with large rewards do not drown out others. 

In the following experiments, we examine \gls{dqn} with several buffering strategies and compare them to oracle baselines and the on-policy multiple-expert approach: a \textbf{(1) Large Buffer} that is akin to saving every sample, a \textbf{(2) Small Buffer}, which inherently prioritizes recent samples,
\textbf{(3) Long-term short-term} that saves experiences in two buffers, one large and one small~\cite{isele2018selective}, and samples from them equally during training to combine data from the entire history with recent samples, and \textbf{(4) Multiple buffers} that keeps a separate buffer for each workload and samples equally from them during training.

Our evaluations use three workload schedules shown in \Cref{fig:scenario_all}. In the on-policy experiments we focused on scenario I where workloads change in a cyclic manner, but here we also consider cases where some workloads might be encountered rarely. Notably, we will find that unlike the on-policy method, the off-policy approach can be sensitive to the schedule.

\begin{figure}[t]
    \includegraphics[trim=0 0.3cm 0 0cm, clip, width=\linewidth]{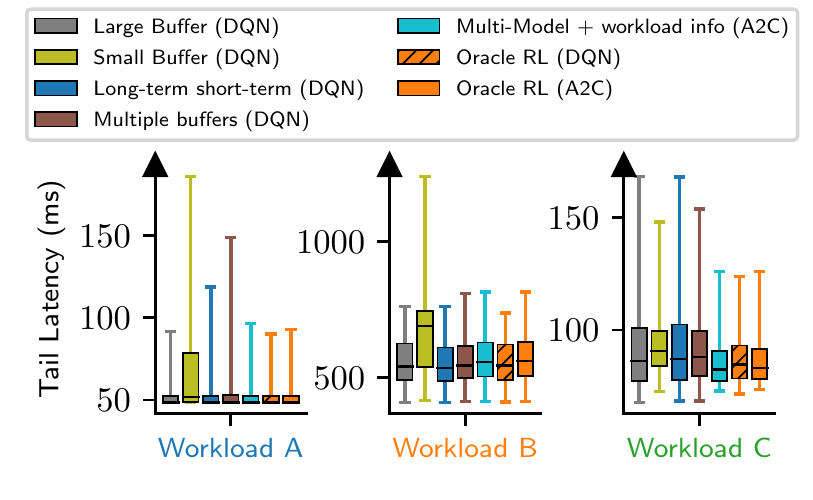}
    \vspace{-20pt}
    \caption{Tail latency distribution after convergence in Scenario I, when $T_{sw}=T_c$ (similar to \Cref{fig:on_policy_CF}). \gls{dqn} with a small buffer forgets past workloads.}
    \label{fig:off_policy_sc0}
    \vspace{-20pt}
\end{figure}

\Cref{fig:off_policy_sc0} demonstrates the results for scenario I. Unsurprisingly, a small buffer does not fare well; it loses experience samples of previous workloads and the agent forgets them, leading to \gls{cf}. Note how in workload C, the \gls{dqn}-based oracle fares better than all other online \gls{dqn} variants. A similar observation arises in \Cref{fig:on_policy_CF}, when using a single model with workload info at a fast workload switching setting. We believe this slight loss is due to shared learning, which can sometimes diminish performance instead of improving it when using a single neural network model for diverse tasks~\cite{Taylor_Stone_2011}. 

\begin{figure}[t]
    \includegraphics[trim=0 0.4cm 0 0.1cm, clip, width=\linewidth]{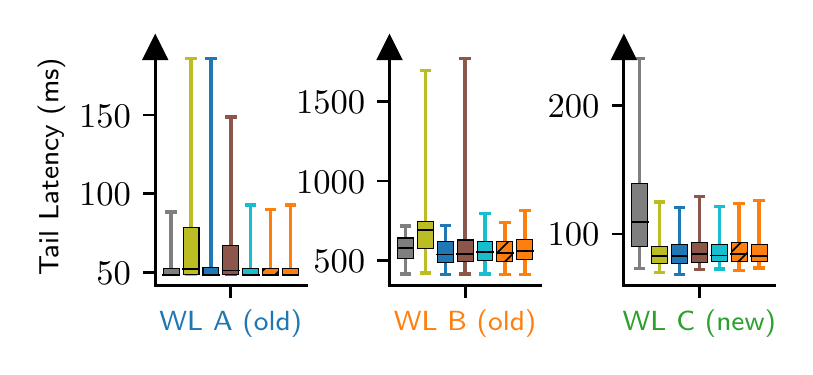}
    \caption{Tail latency distribution after convergence in Scenario II, when $T_{sw}=0.5T_c$. Left and middle plot concern remembering common workloads after training on a new one. Right plot shows performance on a new workload, after training. Large and small buffers fail while long-term short-term uses the best of both to fare best.}
    \label{fig:off_policy_sc1}
\end{figure}

\Cref{fig:off_policy_sc1} shows the same set of schemes in scenario II, where a rare workload does not come up until well into training. We use workload C as this rare workload and A and B as the common ones. The evaluations in this plot show performance after workload C converges (a region shown by the red box in scenario II in \Cref{fig:scenario_all}). In chronological order, the results for workload C show the large buffer struggling; a large buffer will amass a high volume of samples after a long time and a new workload will have a minuscule share of buffer and training samples. The results for workloads A and B on the other hand exhibit the forgetfulness of a small buffer, similar to the previous experiment. Contrary to both methods, long-term short-term performs well. 

\begin{figure}[t]
    \includegraphics[trim=0 0.4cm 0 0.3cm, clip, width=\linewidth]{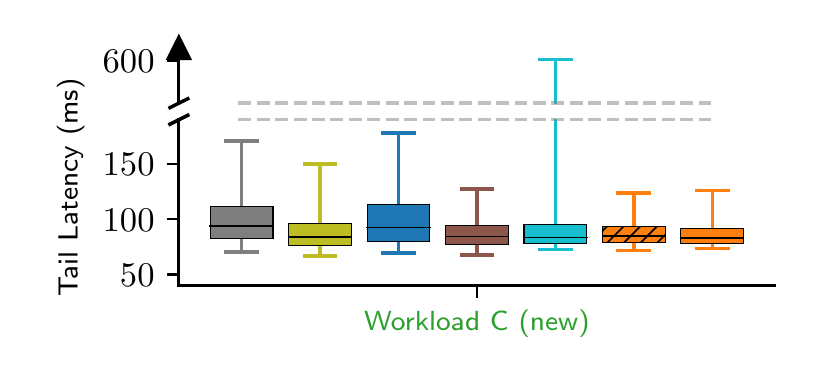}
    \caption{Tail latency distribution after convergence in Scenario III, when $T_{sw}=0.5T_c$. This plot concerns how well an agent remembers a rarely occurring workload. Long-term short-term is outperformed by multiple buffers.}
    \label{fig:off_policy_sc2}
    \vspace{-15pt}
\end{figure}

Finally, \Cref{fig:off_policy_sc2} shows results for scenario III, which demonstrates how well an approach can remember a rarely occurring workload (shown by the red box in \Cref{fig:scenario_all}). Here long-term short-term does not cope well; the small buffer forgets workload C and the large one favors more common ones. Multiple buffers performed well here, but it is worse than long-term short-term in scenario II (\Cref{fig:off_policy_sc1}). Overall, these results show that none of these schemes can be a universal strategy that performs well in all circumstances. By contrast, the on-policy multiple-expert approach is more robust.  

\subsection{Experiments with Real Non-stationarity}
\label{sec:real_results}

So far, we have studied the different dimensions of non-stationarity and their effect on the agents performance, and demonstrated the superiority of the proposed framework in \S\ref{sec:framework_challenges}.
We did so by ``synthetically" switching between several traces. Now we consider traces with natural non-stationarity. We compare our complete framework against three baselines:\footnote{We choose to include the safeguards in baselines as our experiments strongly demonstrate its necessity.} \textbf{1)} \emph{Single-model+workload info} (as in \Cref{fig:on_policy_CF}), \textbf{2)} Single-model DQN from with Long-term short-term buffering strategy (Commonly used in applied \gls{rl} for non-stationarity) and \textbf{3)} Model-Based Change-point Detection (MBCD)~\cite{alegre2021minimum} as a generic environment-detection approach. For MBCD, the default hyperparameters led to hundreds of models being launched. We tuned them so that approximately the same number of models are launched as the multi-model approach.

We evaluate our framework in two scenarios with real non-stationarity traces. First as a departure from the abrupt switches in \S\ref{subsec:lb_exps}, we consider a smoothly changing workload. Second, we study a workload with frequent abrupt changes.

\begin{figure}
    \begin{subfigure}{0.49\linewidth}
        \hspace*{-0.3cm}
        \includegraphics[width=\linewidth]{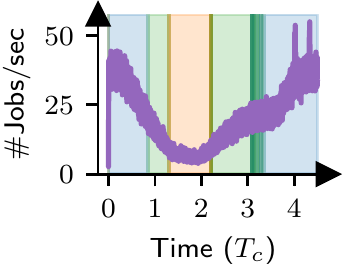}
        \caption{}
        \label{fig:slow_market_detect}
        \vspace{-10pt}
    \end{subfigure}
    \begin{subfigure}{0.49\linewidth}
        \includegraphics[width=\linewidth]{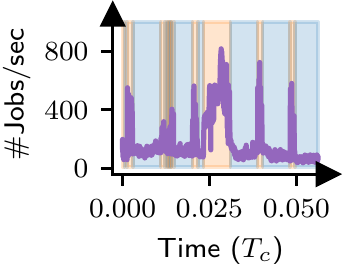}
        \caption{}
        \label{fig:fast_catalog_detect}
        \vspace{-10pt}
    \end{subfigure}
    \caption{Jobs arriving per second in the \emph{Marketplace} \textbf{(a)} and \emph{Catalog} \textbf{(b)} workloads, and how the environment detection module classifies environments. Each background color represents a detected environment, with a total of 3 and 2 environments in \textbf{(a)} and \textbf{(b)}. }
    \label{fig:slow_real_traces}
    \vspace{-12pt}
\end{figure}

\begin{figure}[!t]
    \begin{subfigure}{\linewidth}
        \includegraphics[width=\linewidth]{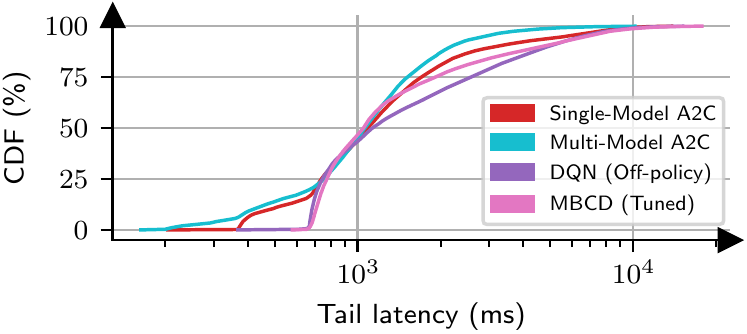}
        \caption{}
        \label{fig:slow_market_cdf}
    \end{subfigure}
    \begin{subfigure}{0.49\linewidth}
        \includegraphics[width=\linewidth]{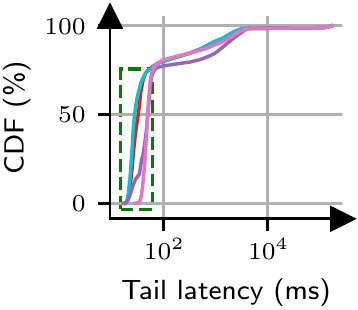}
        \caption{}
        \label{fig:fast_catalog_cdf_full}
        \vspace{-10pt}
    \end{subfigure}
    \begin{subfigure}{0.49\linewidth}
        \includegraphics[width=\linewidth]{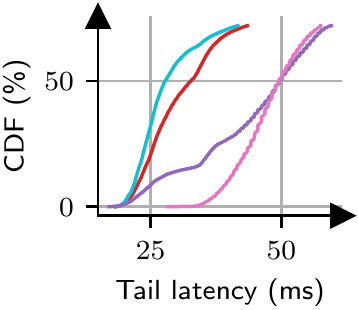}
        \caption{}
        \label{fig:fast_catalog_cdf_partial}
        \vspace{-10pt}
    \end{subfigure}
    \caption{Our framework outperforms prior work and a single-model on-policy baseline in two traces with natural non-stationarity. Tail latency CDF after convergence is plotted, when either a single model is used or multiple, for \emph{Marketplace} \textbf{(a)} and \emph{Catalog} \textbf{(b, c)}. \textbf{(c)} demonstrates the first $70\%$ of the CDF in \textbf{(b)}.}
    \label{fig:fast_real_traces}
    \vspace{-20pt}
\end{figure}

\noindent \textbf{Scenario: Slow and Smooth changes:}
We consider an elongated version of the 24-hour trace, \emph{Marketplace}, visualized in \Cref{fig:slow_market_detect} (and more thoroughly in \Cref{fig:app_workload_time}). This scenario consists a slowly but considerably changing workload. The figure also visualizes how our environment detection module (a Gaussian mixture model fit to the trace) clusters the environment, via the background color in each region. This setting might initially suggest that our framework is unnecessary; a simple \gls{rl} agent might catch up with the slow changes as they come by continually training on fresh interactions. However, as the CDF of tail latency in \Cref{fig:slow_market_cdf} shows (note the logarithmic scale), the framework can fare better $2\times$ and by $22\%$ on average, compared to the single-model A2C baseline.

As explained before, once an agent has finished exploring and enters the exploitation phase, learning is limited to fine-tuning the policy. Even if the workload changes slowly after this point, the single-model A2C baseline will not learn how to interact in this new environment. To illustrate this point, \Cref{fig:slow_market_detect} shows that when convergence is reached around $1 T_c$, the baseline is subject to the green environment with mid-range job arrivals. Per the explanation above, the baseline should perform well with mid-range job arrivals, but sub-optimally at low or high-range job arrival rates. This is clear in \Cref{fig:slow_market_cdf}, where the baseline is closest to our framework in mid-range latencies (Latency is directly correlated with queue sizes, which are correlated with arrival rate). The DQN and MBCD baselines (which uses the off-policy Soft-Actor Critic~\cite{haarnoja2018soft} for learning) fail to outperform even a single-agent A2C. We have evaluated DQN extensively in \S\ref{subsec:lb_exps} and explained why this happens. As for MBCD, a change-point is detected when a learned transition model can not consistently predict the actual evolution of the environment. This approach can work in highly predictable environments with abrupt non-stationary changes. However, real system environments often exhibit high variance, as our case study with real traces shows, and furthermore this trace is changing smoothly, not abruptly. As a result, MBCD fails to classify this environment correctly.

\noindent \textbf{Scenario: Frequent and Abrupt changes:}
A workload that changes frequently and in a considerable range provides an interesting example; the wide range implies several environments in the workload, but the frequent change means that interaction data over a short training span would be a mixture of these different environments, and therefore less likely to cause catastrophic forgetting. Such an example can be viewed in \Cref{fig:fast_catalog_detect}, which shows a sample of the \emph{Catalog} trace. Note the frequency of changes with respect to the convergence time. The figure also shows how the environment detection module (a Gaussian mixture model fit to the trace) can detect the `idle' (low) and `rushed' (high) regions.

Despite the setting, our framework still trains better performing agents than baselines, as evident in \Cref{fig:fast_catalog_cdf_full,fig:fast_catalog_cdf_partial}. The high range of the workload causes drastically different latency patterns to emerge and the tail latency CDF spans a very wide range, occluding differences in the idle portions of the trace. The inferiority of single-model A2C is due to this vast range; the reward of the agent is tail latency and during rushed portions it observes rewards several magnitudes larger than rewards in the idle phase. The underlying neural networks obey the loss function and over-prioritize learning for the rushed portion. As a result, the baseline performs well in the rushed regions with high latency and rewards, but underperforms in the simpler idle regions with low rewards. Our framework naturally isolates these two regions and avoids a mixture of incompatible loss scales. Since changes are abrupt, one might hope that MBCD can isolate these two regions. However, in this real trace changes are also quick. MBCD requires the environment to remain in a workload after a change for a minimum amount of time, and states that for a generic change-point detection, this assumption is critical. Since this trace violates that assumption, MBCD can't distinguish high and low load regions, and launches new models at random.

\section{Case Study: Adaptive Bit Rate}
\label{sec:abr}

We now apply the framework from \S\ref{sec:framework_challenges} to the \gls{abr} environment.
%
\label{subsec:abr_setup}
In this environment, a server streams video to a client.
The video is divided into $T$ second chunks and each chunk is encoded at several bitrates
of varying quality and file size.
For each chunk, an \gls{abr} algorithm decides which bitrate to send.
At the client-side, a playback buffer aggregates received chunks and displays the video to the
client, draining the buffer at a constant rate.
Formally, at step $t$ and when the clients buffer is $b_t$, the server streams the next video
chunk with size $S_t$ and bandwidth $c_t$.
If the chunk arrives after the buffer depletes, the video pauses and the client observes a
\emph{rebuffer event}. Generally, the next buffer will be:
$$b_{t+1} = max(0, b_t-\frac{S_t}{c_t})+T$$
The client only requests chunks when the buffer falls below threshold $M_b$.
There are three goals in this environment; \textbf{avoid rebuffer events},
\textbf{stream in high quality}, and \textbf{maintain a stable quality}.
These goals are conflicting in nature, and systems in practice optimize a linear combination of
them, coined \gls{qoe}~\cite{yin2015control}.\footnote{There are numerous methods for defining
\gls{qoe} metrics~\cite{mao2017neural, spiteri2020bola, yin2015control}. The choice of
\gls{qoe} is orthogonal to our discussion.}
Assuming the bitrate of step $t$ was $q_t$, the \gls{qoe} is:
$$QoE_t = q_t - |q_{t}-q_{t-1}| - \mu \cdot max(0, \frac{S_t}{c_t}-b_t)$$
To learn a policy for this problem, we have to define the observations, reward and actions.
Each step in the environment is a video chunk and the action is what bitrate is sent.
The reward is the \gls{qoe} of that chunk.
The observation $o_t$ is comprised of \textbf{(a)} download times and bandwidths in the last
$K=9$ chunks, \textbf{(b)} buffer occupancy before sending chunk $t$, \textbf{(c)} how many
chunks are left in the video, \textbf{(d)} the previous chunk's bitrate (previous action) and
\textbf{(e)} available bitrates for chunk $t$.
For more details on the \gls{abr} environment and training setup, see \S\ref{app:abr_train}.

\begin{table}
\small
\centering
\begin{tabular}{r | c  c c }

\textbf{Group} & \textbf{Average BW}  & \textbf{Individual variance}  & \textbf{Diversity} \\
\toprule
\midrule
UG 1 & Low & High & Medium \\
UG 2 & High & Low & High\\
UG 3 & Medium & Medium & Low\\
UG 4 & Medium-Low & Medium & Low\\
UG 5 & Medium & High & Very High\\

\end{tabular}
\caption{High level description of User Groups. Refer to \S\ref{app:user_bw} in the appendix for further details.}
\label{table:abr_groups}
\vspace{-5pt}
\end{table}

The network path distribution of active streaming sessions is the non-stationary element in this problem.
A server responds to and shares a single learned policy for all clients, but the users of the service will
change over time (e.g., children in the morning and adults at night at a daily scale, and events such as
streaming sports or election news at a weekly/yearly scale).
With them, the network path distribution of active sessions changes.
The networks themselves can also change: e.g., cellular use in mornings and broadband at night.
Further, changes such as new network infrastructure and protocols affect network sessions.
To model this, we switch between one of five bandwidth distributions from a generative model, similar to
scenario I (see \Cref{fig:scenario_all}).
\Cref{table:abr_groups} provides a high level description of how these groups differ.
For details about these distributions and sample bandwidth traces, see \S\ref{app:user_bw}.
%

\subsection{Applying our framework to non-stationary \gls{abr}}
\label{subsec:abr_exp}

\begin{figure}[t]
    \includegraphics[trim=0 0.4cm 0 0.4cm, clip, width=\linewidth]{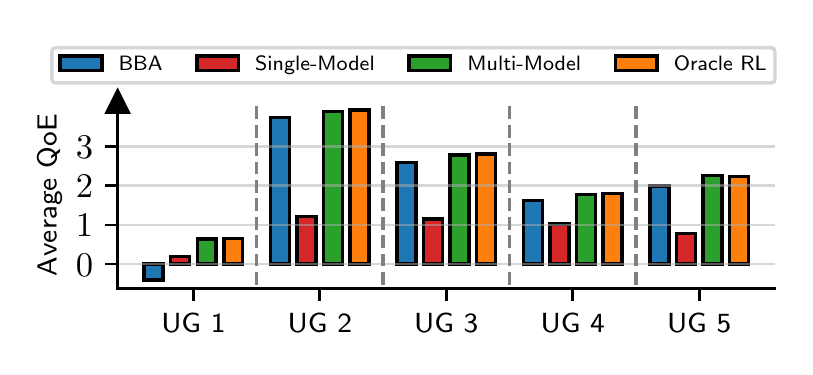}
    \caption{Average \gls{qoe} after convergence in Scenario I when $T_{sw}=T_c$, per different user groups. \emph{UG X} refers to \emph{User Group X}. Higher is better.}
    \label{fig:abr_multi}
    \vspace{-18pt}
\end{figure}

In our safe guard (explained in detail in \S\ref{subsec:abr_tw}), the agent starts off with control over a limited number of chunks, where wrong decisions do not damage \gls{qoe} and use a default policy (linear BBA~\cite{bba}) elsewhere. Gradually, we allow control over riskier decisions until the agent converges. To learn risky scenarios without direct control, we use domain knowledge about buffer dynamics to provide the agent with an alternate risky experience. We take care to not bias the \gls{rl} training methodology.

To detect user distributions (environments), we use the same trained classifier approach as before. This classifier looks at the average bandwidth and it's variability at two time scales. We use this classifier to dynamically control the exploration. \gls{a2c} with multiple experts proved reasonably robust in the previous case study and therefore we use this approach here.

\Cref{fig:abr_multi} compares \gls{bba}, a single expert and an oracle baseline against the multiple expert approach. We consider \gls{bba} as a non-learning based alternative. As observed, multiple experts' results are similar to the oracle, which means the multiple experts approach has completely side-tracked non-stationarity. Also, the multiple experts approach achieves favorable results compared to \gls{bba}, while the single expert baseline does not. This implies that without dealing with non-stationarity, utilizing learned policies via \gls{rl} will not be practical. By considering this important aspect, evident in many real-world problem, one can truly utilize the benefits of \gls{rl}. For a breakdown and interpretation of these \gls{qoe} results, see \S\ref{app:abr_qoe_breakdown}.

\section{Related Work}
\label{sec:related_works}

Herein we explore related works other than those considered in \S\ref{sec:non_stat_approach}.
\textbf{Non-stationarity in \gls{rl}:} Non-stationarity has been explored in \gls{rl} in various contexts with varying assumptions, but a general solution has not been proposed~\cite{khetarpal2020towards}.
One class of methods train meta models prior to deployment, and use few-shot learning to adapt after deployment~\cite{nagabandi2019deep, alshedivat2018continuous, nagabandi2019learning}.
However, such methods require access to the environments before deployment, which we do not assume. A body of work proposes using signals from the environment to drive exploration, such as novelty of a visited state~\cite{pathak2017curiosity}. Such methods can potentially be integrated to our framework to help with exploration.
For a comprehensive review of non-stationary \gls{rl}, refer to~\cite{khetarpal2020towards}.

\textbf{Safety in RL:} Safe \gls{rl} takes on many definitions~\cite{garcia2015comprehensive}. Our focus is on avoiding disastrous performance outcomes in a live system while exploring or exploiting.
\cite{rotman2020online} use several forecast signals and revert to a default policy whenever the agent is prone to mistakes. 
\cite{dalal2018safe} use logged data to learn when a set of constraints can be violated, and disallow the agent from taking such actions in deployment. 
Several approaches attempt to keep safety by assuming regularity~\cite{achiam2017constrained} or smoothness~\cite{chandak2020safe} in the system. Another model-based RL approach uses active learning and Gaussian Processes for exploration and avoiding safety violations~\cite{cowen2022samba}.
For a full review of safety in RL, refer to ~\cite{garcia2015comprehensive}.
Using logged interaction data from a deployed policy, one could bootstrap a safe policy for deployment~\cite{thomas2015safe,garcia2015comprehensive}. 
However, even assuming logged data provides full coverage over the state space, such methods are prone to distributional shifts caused by train/test mismatch in environments~\cite{levine2020offline}.

\textbf{\gls{cf} in RL:} Learned models are prone to catastrophically forgetting their previous knowledge when training sequentially on new information~\cite{MCCLOSKEY1989109, PARISI201954}. Recently, interest has piqued concerning \gls{cf} in \gls{rl} problems. Three general approaches exist for mitigation~\cite{PARISI201954}: (1)~regularizing model parameters so that sequential training does not cause memory loss~\cite{kirkpatrick2017overcoming, kaplanis2018continual}; (2) selectively training parameters for each task and expanding when necessary~\cite{rusu2016progressive}; (3) using experience replay or rehearsal mechanisms to refresh previously attained knowledge~\cite{Atkinson_2021, rolnick2019experience, isele2018selective}; or combinations of these techniques~\cite{schwarz2018progress}. For a full review of these approaches, refer to~\cite{PARISI201954}.

\section{Conclusion}
We investigated the challenges of online RL in non-stationary systems environments, and proposed a framework to address these challenges. Our work shows that an RL agent must be augmented with several components in order to learn robustly while minimally impacting system performance. In particular, we propose an environment detector to control exploration and select an appropriate model for each environment; and we also propose a safety monitor and default policy that protect the system when it enters an unsafe condition. 
Our evaluation on two systems problems, straggler mitigation and bitrate adaptation, shows that applying our framework leads to policies that perform well in each environment and remember what they have learned. 

\bibliography{paper}
\bibliographystyle{mlsys2023}

\clearpage
\appendix
\begin{appendices}
\appendix
\section{\gls{rl} algorithms}
\subsection{\gls{a2c}}
\label{app:a2c}
The main idea in \gls{a2c} is to calculate the gradient of the expected returns with respect to a policy $\pi_\theta$, characterized by parameters $\theta$ and based on sampled interactions using $\pi_\theta$. 

$$\nabla_\theta \mathbb{E}_{\pi_\theta} [\sum_{t=0}^H \gamma^t r_t] = \mathbb{E}_{\pi_\theta} [\nabla_\theta \log \pi(s_t, a_t; \theta) R_t]$$

$R_t$ (return in state $s_t$) is a high-variance random variable. To decrease the variance and improve training speed and stability, a baseline can be deducted from the return, without biasing the gradient. The used baseline is the expected return when using $\pi_\theta$ ($V_{\pi_\theta}(s_t)=\mathbb{E}_{\pi_\theta}[R_t]$), and the result of decreasing this baseline is called the advantage. Finally, using stochastic gradient ascent we have:
$$\theta \leftarrow \theta + \alpha_\pi \mathbb{E}_{\pi_\theta} [\nabla_\theta \log \pi(s_t, a_t; \theta) (R_t-V_{\pi_\theta}(s_t))]$$
We further improve the variance of \gls{a2c} by employing \gls{gae}~\cite{schulman2018highdimensional}.
We also use entropy regulation~\cite{williams1991function, mnih2016asynchronous} to incentivize exploration during training. The entropy term decays from an initial value to zero (full exploitation), as non-zero values lead to sub-optimal policies~\cite{ahmed2019understanding}. 
To estimate the expected return, we can train a separate critic with parameters $\phi$ using mean squared error loss and returns as targets:
$$\phi \leftarrow \phi - \alpha_v \mathbb{E}_{\pi_\theta} [\nabla_\phi (V^{\pi_\theta}(s_t; \phi) - R_t)^2]$$

\subsection{\gls{dqn}}
\label{app:dqn}
The optimal Q-value $Q^*(s_t, a_t)$ is the value of picking action $a_t$ in state $s_t$, when all subsequent actions come from an optimal policy. We can learn $Q^*$ by repeatedly applying the optimal Bellman operator to any starting estimate:
$$Q^*(s_t, a_t) = r(s_t, a_t) + \gamma \max_{a_{t+1} \in \mathcal{A}} [Q^*(s_{t+1}, a_{t+1})]$$
This operator holds, regardless of what policy created the sample. Also note that the optimal policy is the action with the highest Q-value, also known as the greedy policy. So \gls{dqn}'s training procedure starts by picking random actions and using the samples to train a Q-network. As training goes forward and the Q-network's estimates of the optimal Q-values are more precise, we gradually shift the policy from random to optimal.

We use \gls{ddqn} in this paper, which uses the same principle as \gls{dqn} but uses techniques to reduce overestimation and improve stability~\cite{doubledqn}. We also apply soft target network updates (also known as Polyak averaging) to increase stability~\cite{lillicrap2019continuous}.

\clearpage
\section{Straggler Mitigation}
\subsection{Workloads}
\label{app:workload}

\begin{figure}[t!]
    \includegraphics[width=\linewidth]{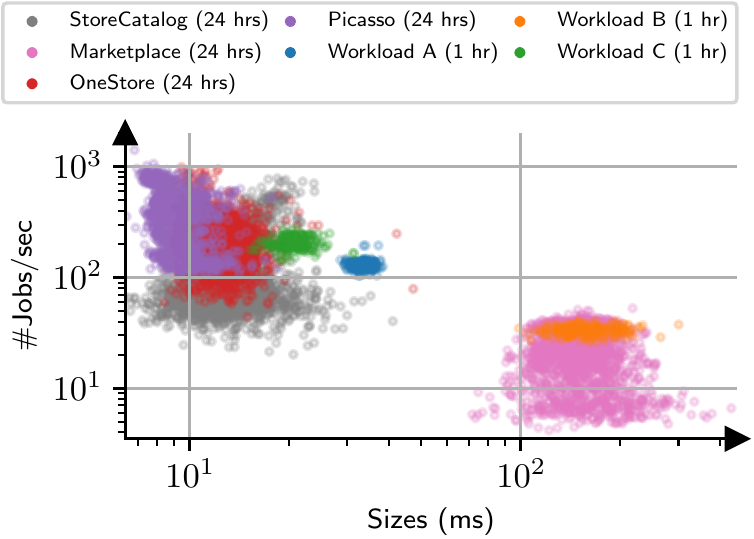}
    \caption{Arrival rates and sizes, for six workloads from five traces (\emph{Workload B} is from the \emph{Marketplace} trace). Each point shows five-second averages. Note the $\log_{10}$ scale on both axes. 
    \label{fig:app_offline_visual}
    }
    \vspace{-8pt}
\end{figure}

We have a trace from a production web framework cluster at AnonCo, collected from a single day in February 2018. The framework processes high-level web requests and breaks them down into subrequests that are issued to various backend services (e.g., to get product details, recommended items, billing information, etc.). These traces exhibit substantial variance, both in short durations and in longer ones. We use several of the traces from the different backend services as workloads in our straggler mitigation experiments:
\begin{itemize}
    \item \textbf{Picasso} from \emph{picassoPageObjectServiceV8}; a full-day trace. Used in evaluating offline training.
    \item \textbf{OneStore} from \emph{oneStoreRecommendationService}; a full-day trace. Used in evaluating offline training.
    \item \textbf{StoreCatalog} from \emph{oneStoreDisplayCatalogService}; a full-day trace. Used in evaluating offline training.
    \item \textbf{Marketplace} from \emph{Marketplace Collections/Entitlements Service}; a full-day trace. Used in evaluating offline training.
    \item \textbf{Workload A} from \emph{TreatmentAssignmentService}; a 1-hour trace. Used in experiments with non-stationarity.
    \item \textbf{Workload B} from \emph{Marketplace Collections/Entitlements Service}; a 1-hour trace. Used in experiments with non-stationarity.
    \item \textbf{Workload C} from \emph{Marketplace Catalog}; a 1-hour trace. Used in experiments with non-stationarity.
\end{itemize}
These workloads are visualized as a scatterplot in \Cref{fig:app_offline_visual} and time series in \Cref{fig:app_workload_time}. Shaded lines denote each attribute of a workload, namely job sizes and arrival rates, in 500ms time windows. Solid color lines are Exponentially Weighted Moving Averages of the shaded lines.

\subsection{Extended results}
\label{app:lb_more_results}

\subsubsection{Offline training}
\label{app:lb_offline_train}

\begin{figure*}
  \includegraphics[width=\linewidth]{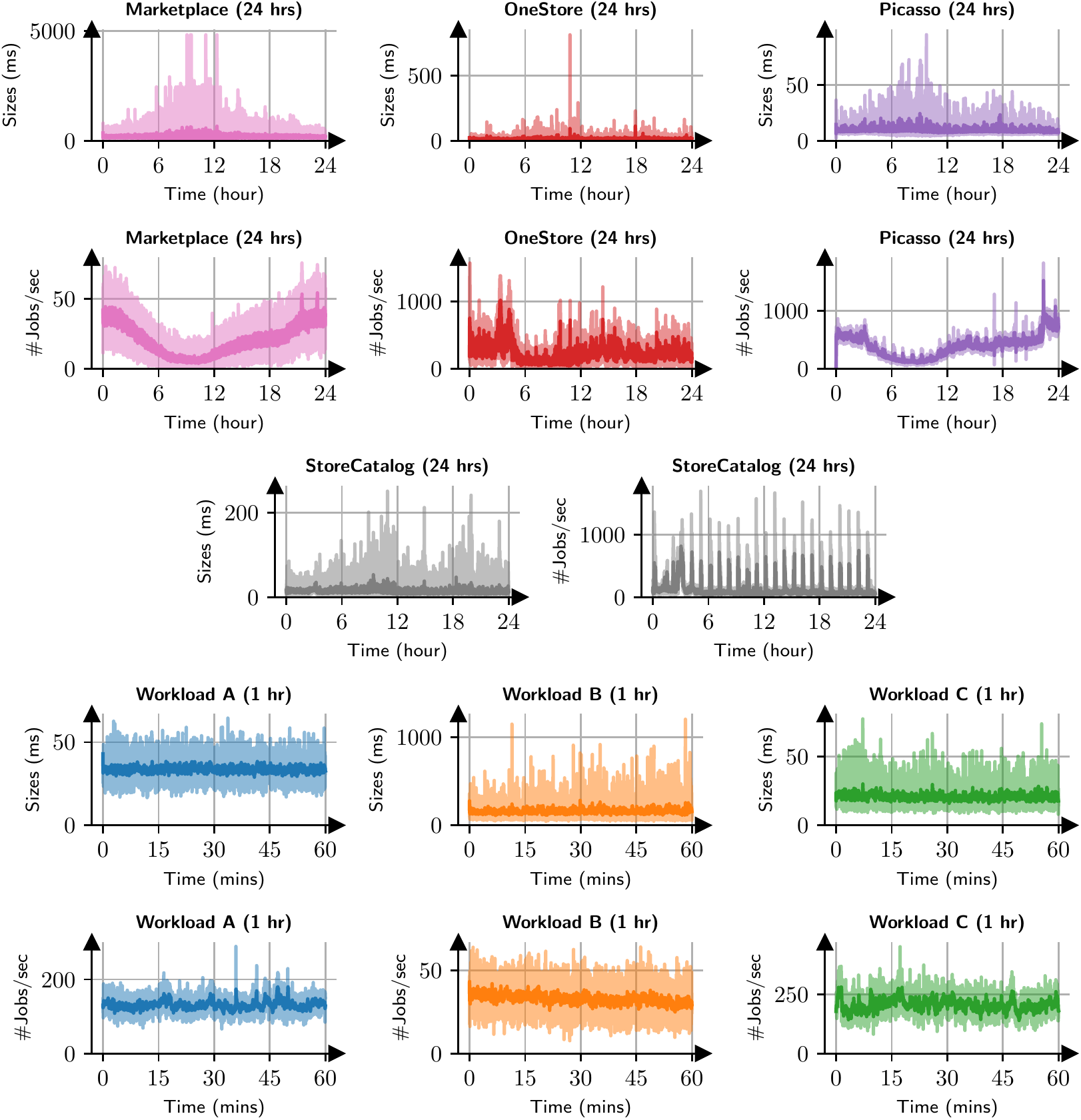}
  \caption{Visual depiction of workloads through time.}
  \label{fig:app_workload_time}
\end{figure*}

Suppose we obtain a faithful simulator for a system, use it to train a policy, and then deploy the policy in the live system. 
We can compare such a pre-trained offline policy to a hypothetical ``oracle" RL policy, which knows the workload that will appear in each interval, and uses a policy trained specifically for that workload ahead of time. 
Concretely, we train an \gls{rl} agent using a training workload and the on-policy \gls{a2c} algorithm. We deploy the final policy to an environment with a test workload.
Train and test workloads are chosen from three day-long traces (\emph{Marketplace Collections}, \emph{OneStore} and \emph{Picasso}), as depicted in \Cref{fig:app_workload_time}. 

\begin{figure}[t]
    \includegraphics[trim=0 0.4cm 0 0.3cm, clip, width=\linewidth]{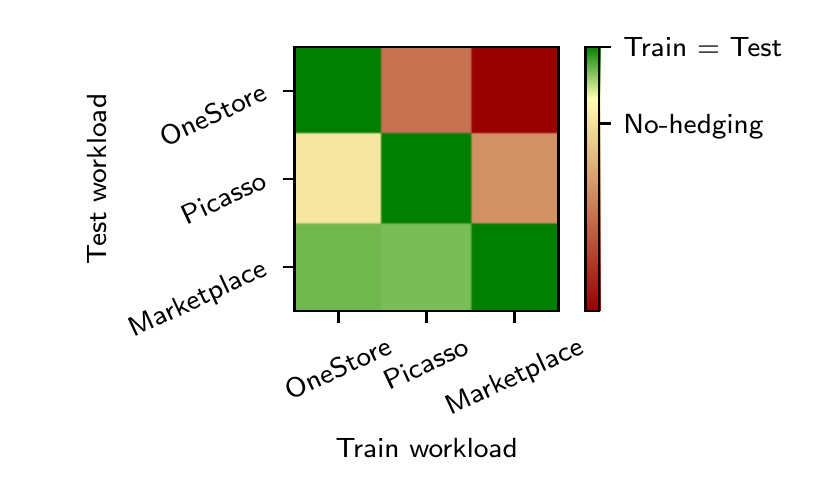}
    \caption{Average tail latency when testing offline-trained models on different workloads. Offline models' test performance degrades if the workload differs from the training workload, with stronger degradation for more dissimilar workloads.}
    \label{fig:app_offline_heatmap}
\end{figure}

\Cref{fig:app_offline_heatmap} visualizes the performance of a model based on it's train/test workload. This heatmap normalizes the average tail latency with the performance of a no-hedging agent, and the agent with matching train/test workloads. Note the clear degradation in performance when the test/train workloads do not match, even for similar workloads such as \emph{OneStore} and \emph{Picasso}. Offline training, even under the favourable conditions of a perfect simulator and stable on-policy training is not viable.
This is caused by the differences between the training environment and deployment environment~\cite{christiano2016transfer, rusu2018simtoreal}.
Although one can try to create representative training datasets, it is difficult, if not impossible, to anticipate and capture every behavior that can occur in a live system~\cite{yan2020learning}.

\begin{figure}[t]
    \includegraphics[trim=0 0.1cm 0 0.3cm, clip, width=\linewidth]{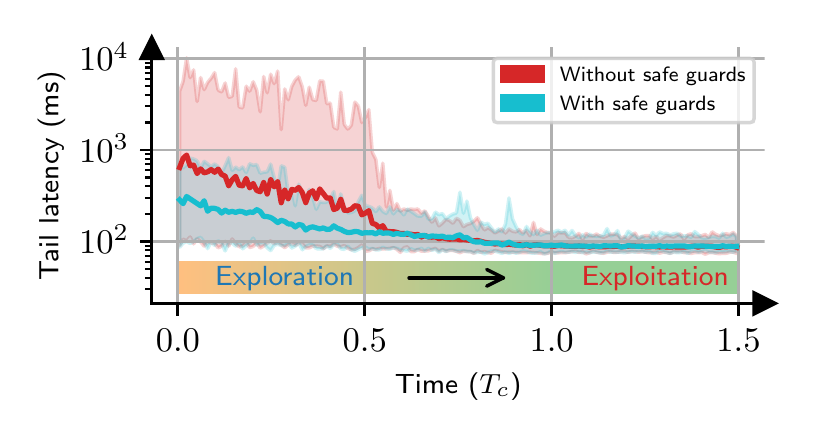}
    \vspace{-20pt}
    \caption{Tail latency over time in workload C, when using a safe guard to reduce performance loss in exploration, and when not. Shaded regions denote tail latency min-to-max in 3 random seeds.}
    \label{fig:on_policy_tw}
    \vspace{-12pt}
\end{figure}

\begin{figure}[t]
    \includegraphics[trim=0 0.3cm 0 0.3cm, clip, width=\linewidth]{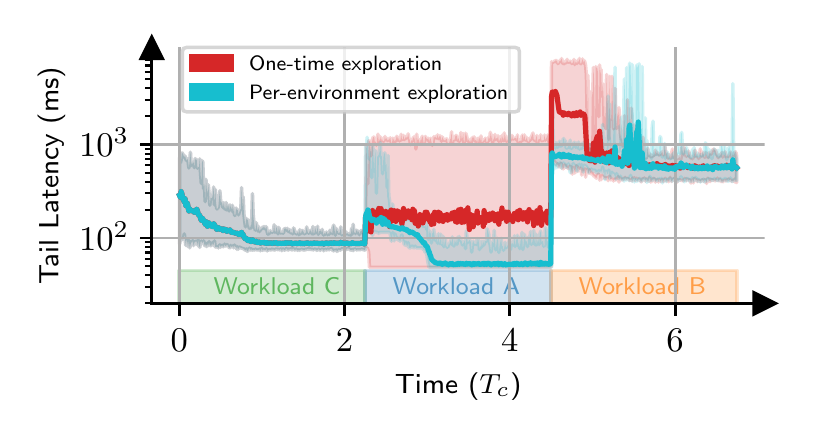}
    \caption{Tail latency across time, when we start with workload C. Shaded regions denote min-to-max of tail latency across 3 random seeds. Each workload is given enough time to converge ($T_{sw} = 2.25 T_c$).}
    \label{fig:app_on_policy_reset_timeseries}
\end{figure}

\begin{figure}[t]
    \includegraphics[trim=0 0.4cm 0 0.7cm, clip, width=\linewidth]{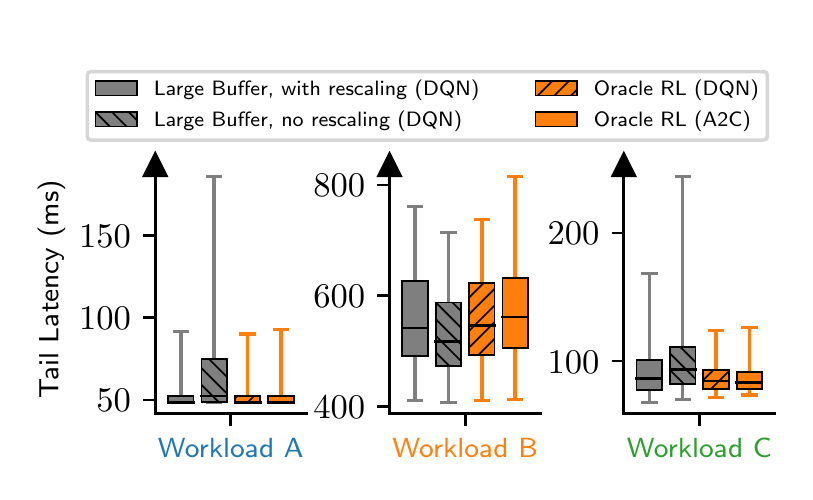}
    \caption{Tail latency distribution after convergence in Scenario I, when $T_{sw}=T_c$ (similar to \Cref{fig:on_policy_CF}). Without considering rewards scales, \gls{dqn} favors workloads with high magnitude rewards at the expense of others.}
    \label{fig:app_off_policy_sc0_scaling}
\end{figure}

\subsubsection{Safeguards bound performance loss in exploration}
\label{app:safe_explore}
As discussed before, training online in a live system requires online exploration, which leads to transient loss in performance. While our main priority is to achieve strong eventual performance, bounding transient degradation while exploring is a secondary but important goal. We achieve this using safeguard policies, as explained in \S\ref{sec:framework_challenges}. 
\Cref{fig:on_policy_tw} demonstrates the effect. Without safeguards, latencies can be an order of magnitude higher than the worst case performance with safeguards. 
 
We observed that occasionally training sessions without safeguards failed to learn at all and queues grew beyond tens of thousands of jobs. \Cref{fig:challenges_time_series} showed an example of such a failure mode. This occurs because when queues exceed a certain bound, all jobs will take longer than the maximum hedging timeout (300~ms) and get duplicated. In such a situation, the only way to drain the queues is to disable hedging entirely. But while the agent tries to learn this behavior, the queues continue to grow and the system repeatedly reaches states that the agent has never seen before. Hence, the agent is not able to spend enough time exploring the same region of the state space to learn a stabilizing policy and a vicious cycle forms. 

\subsubsection{Per-environment exploration}
\label{app:lb_more_exp}
\Cref{fig:app_on_policy_reset_timeseries} demonstrates the effect of exploring once for each workload. This plot shows the tail latency over the course of the experiment. We see that while one-time exploration (for $T_c$ time) fails to handle new workloads (e.g., Workload A), per-environment exploration converges to a good policy for each workload. One-time exploration is clearly not a viable approach.

\subsubsection{DQN scaling}
\label{app:lb_off_scaling}

A further nuance with the off-policy approach, is reward scaling. Since the Q-value network will train on samples from multiple workloads with different magnitudes (latency in one workload can be $10~ms$ while $500~ms$ in another), workloads with larger rewards will be overemphasized in the L2 loss in training. Therefore, the rewards must be normalized but done so consistently. \Cref{fig:app_off_policy_sc0_scaling} shows the effect; If scaling is not performed, workloads with smaller rewards (latencies), such as workloads A and C are sacrificed for superior performance in workloads with large rewards. Besides manual normalization, methods exist for automatic adaptation to reward scales~\cite{vanhasselt2016learning,hessel2018multitask}.

\subsection{Training and environment setup}
\label{app:lb_train}

Our implementations of \gls{a2c} and \gls{dqn} use the Pytorch~\cite{pytorch} library. 
\Cref{table:lb_train} is a comprehensive list of all hyperparameters used in training and the environment.

\clearpage
\begin{table*}
\small
\centering
\begin{tabular}{c l l }
\toprule
\textbf{Group} & \textbf{Hyperparameter} & \textbf{Value} \\
\cmidrule{2-3}

\multirow{16}{*}{Neural network} & \multirow{2}{*}{Hidden layers} & $\phi$ network: (16, 8) \\
\cmidrule{3-3}
& & $\rho$ network: (16, 8) \\
\cmidrule{2-3}
& Hidden layer activation function & Relu \\
\cmidrule{2-3}
& \multirow{4}{*}{Output layer activation function} & \gls{a2c} actor: Softmax \\
\cmidrule{3-3}
& & \gls{a2c} critic: Identity mapping\\
\cmidrule{3-3}
& & \gls{dqn}: Identity mapping\\
\cmidrule{2-3}
& Optimizer & Adam \cite{kingma2017adam} \\
\cmidrule{2-3}
& Learning rate & 0.001 \\
\cmidrule{2-3}
& $\beta_1$ & 0.9 \\
\cmidrule{2-3}
& $\beta_2$ & 0.999 \\
\cmidrule{2-3}
& $\epsilon$ & $10^{-8}$ \\
\cmidrule{2-3}
& Weight decay & $10^{-4}$ \\
\midrule

\multirow{6}{*}{\gls{rl} training (general)} & Episode lengths & 128 \\
\cmidrule{2-3}
& Epochs to convergence ($T_c)$ & 6000 (728000 samples) \\
\cmidrule{2-3}
& Random seeds & 3 \\
\cmidrule{2-3}
& $\gamma$ & 0.9 \\
\midrule

\multirow{2}{*}{\gls{a2c} training} & Entropy schedule & 0.1 to 0 in 5000 epochs \\
\cmidrule{2-3}
& $\lambda$ (for \gls{gae}) & 0.95 \\
\midrule

\multirow{8}{*}{\gls{dqn} training} & Initial fully random period & 1000 epochs \\
\cmidrule{2-3}
& $\epsilon$-greedy schedule & 1 to 0 in 5000 epochs \\
\cmidrule{2-3}
& Polyak $\alpha$ & 0.01 \\
\cmidrule{2-3}
& Buffer size in Last $N$ & $10^6$ \\
\cmidrule{2-3}
& Buffer size in Long-term short-term& $10^6$ (short) \\
\cmidrule{2-3}
& Buffer size in Multiple buffers & $10^6$ for each \\
\midrule

\multirow{8}{*}{MBCD baseline} & h & 300000 (default was 100/300) \\
\cmidrule{2-3}
& max\_std & 3 (default was 0.5) \\
\cmidrule{2-3}
& N (Ensemble size) & 5 \\
\cmidrule{2-3}
& NN Hidden layers & (64, 64, 64) \\
\cmidrule{2-3}
& Hidden layer activation function & Relu \\
\cmidrule{2-3}
& Output layer activation function & Identity mapping \\
\midrule

\multirow{9}{*}{Environment} & Number of servers $n$ & 10 \\
\cmidrule{2-3}
& Number of actions & 7 \\
\cmidrule{2-3}
& Timeouts of actions (millisecs) & $\{3, 10, 30, 60, 100, 300, \infty(\text{no-hedging})\}$ \\
\cmidrule{2-3}
& Action time window & 500 ms \\
\cmidrule{2-3}
& (Slow down rate $k$, probability $p$) & (10, 10\%) \\
\cmidrule{2-3}
& Observation workload aggregation steps ($m$) & 4 \\
\cmidrule{2-3}
& Safe guard unsafety upper limit & 50 jobs \\
\cmidrule{2-3}
& Safe guard return to safety lower limit & 3 jobs \\
\midrule
\bottomrule
\end{tabular}
\caption{Training setup and hyperparameters for straggler mitigation experiments.}
\label{table:lb_train}
\end{table*}

\clearpage
\section{Adaptive Bit Rate}
\subsection{User Groups}
\label{app:user_bw}

To create various bandwidth distributions for our experiments, we use a generative process. This process was designed to cover three important characteristics in bandwidth traces: \textbf{(a)} The average bandwidth the whole distribution observes. This represents how fast the users' networks are in total. \textbf{(b)} The variation in average bandwidth among users. This represents how similar or different networks of a group's users are. \textbf{(c)} The variation in the bandwidth trace for each user. This represents how stable a user's network is at providing bandwidth.

The user groups we generate provide a good coverage over these three characteristics:
\begin{itemize}
    \item \textbf{User group 1} is a low bandwidth group, with medium diversity among users and high variance in each user trace.
    \item \textbf{User group 2} is a high bandwidth group, with high diversity among users and low variance in each user trace.
    \item \textbf{User group 3} is a medium bandwidth group, with low diversity among users and medium variance in each user trace.
    \item \textbf{User group 4} is a medium to low bandwidth group, with low diversity among users and medium variance in each user trace.
    \item \textbf{User group 5} is a medium bandwidth group, with very high diversity among users and high variance in each user trace.
\end{itemize}

\begin{figure*}
  \includegraphics[width=\linewidth]{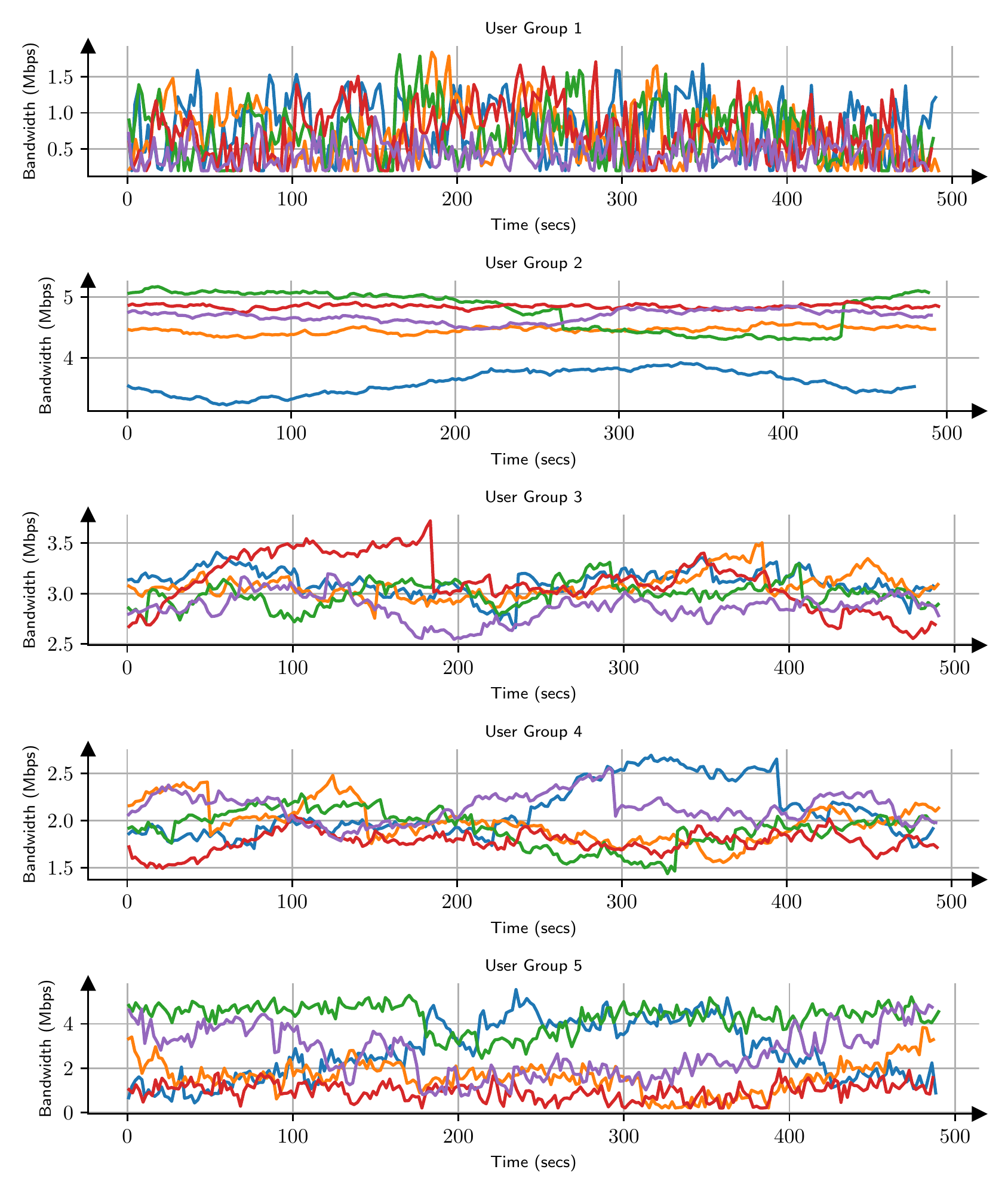}
  \caption{Visual depiction of user group bandwidths through time. Each plot has 5 samples.}
  \label{fig:abr_trace_time}
\end{figure*}

Sample traces for these groups can be observed at \Cref{fig:abr_trace_time}.

The generative process is a mixture of a Markov chain and an Ornstein–Uhlenbeck process~\cite{uhlorn}. The Markov chain generates a coarse bandwidth trace, which prior work shows is an appropriate model for TCP throughput~\cite{cs2p}. The Ornstein–Uhlenbeck process, which is a mean-reverting variant of Brownian motion adds a random walk to the coarse output of the Markov chain. Contrary to previous work~\cite{oboe}, we use a mean reverting random walk instead of i.i.d. Gaussian noise to model smooth changes in throughput as well. We use a uniform initial distribution for the Markov chain. Overall, the degrees of freedom in this process are: \textbf{(a)} the state space of the Markov chain, \textbf{(b)} the transition kernel of the Markov chain, \textbf{(c)} the scale and dissipation rate of the Ornstein–Uhlenbeck process. 

\subsection{Safe guard policy}
\label{subsec:abr_tw}

\begin{figure}[t!]
    \begin{subfigure}{\linewidth}
        \includegraphics[width=\linewidth]{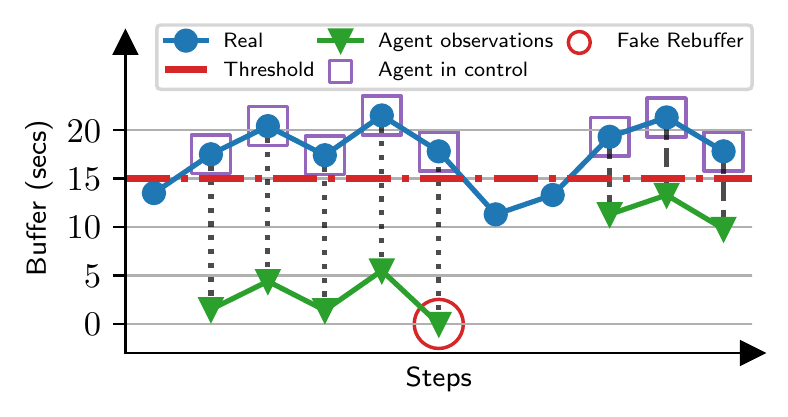}
        \caption{}
        \label{fig:abr_tw_buff_ex}
    \end{subfigure}
    \begin{subfigure}{\linewidth}
        \includegraphics[trim=0 0 0 0.2cm, clip, width=\linewidth]{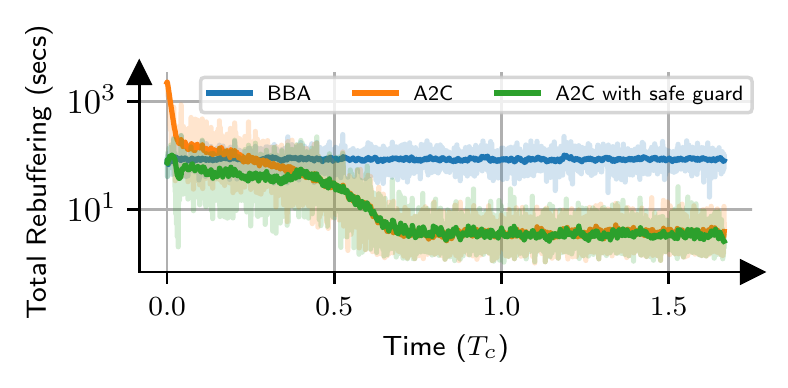}
        \caption{}
        \label{fig:abr_tw_time_series}
    \end{subfigure}
    \caption{\textbf{(a)} An example of how the safe guard operates. \textbf{(b)} Total rebuffering across time, for three variants, under user group 1. Note that the y-axis is logarithmic, due to wide range.}
\end{figure}

Transient performance is important in the video streaming setting.
If the \gls{rl} agent's exploration phase causes significant rebuffering, user satisfaction will drop.
\Cref{fig:abr_tw_time_series} shows how much rebuffering is caused by \gls{a2c} while training.
In initial exploration stages, \gls{a2c} rebuffers as long as the video length, while by the end rebuffers
are in the range of a few seconds.
Thus, we design a safe-guard policy for \gls{abr} training, similar to the one introduced in \S\ref{subsec:lb_setup}.

A natural idea is to gradually give control to the agent, i.e., start it off with control over a limited number of chunks and,
after convergence, give full control.
The primary goal is to avoid rebuffering, we so can give control to the agent when rebuffering is less likely (e.g., when
buffer occupancy is high).
Hence, we give control to the agent whenever the buffer occupancy is above a threshold (the red line in
\Cref{fig:abr_tw_buff_ex}).
For buffer levels below the threshold, we use a reliable but sub-optimal algorithm like \gls{bba}~\cite{bba}.
We then slowly anneal the threshold to promote the agent to full control.
This creates an issue: the agent's exploration phase is biased towards high buffer states, and the agent fails to learn how
to behave in risky low buffer situations.
To mitigate this, we fool the agent into thinking the buffer is at a lower value. Note that since the buffer dynamics in \gls{abr} are simple, this is possible in a real system. Whenever the agent gains control from the safe guard, a buffer value is chosen uniformly at random from the actual buffer and 0. The agent's following observations will be based on this initial buffer value. \Cref{fig:abr_tw_buff_ex} provides a visual example. This ruse continues until the real buffer falls below the threshold and the safe guard takes over. 

\Cref{fig:abr_tw_time_series} compares \gls{a2c} with Fake-Replay, \gls{a2c} and \gls{bba}, while the \gls{rl} agents are training over a challenging bandwidth distribution. As observed, \gls{a2c} with no safeguards begins training with significant rebuffering, but over time learns to rebuffer even less than \gls{bba}. \gls{a2c} with Fake-Replay does not rebuffer overwhelmingly like \gls{a2c}, and its final performance is unaffected by the safe-guard.

\begin{figure}[t!]
    \includegraphics[width=\linewidth]{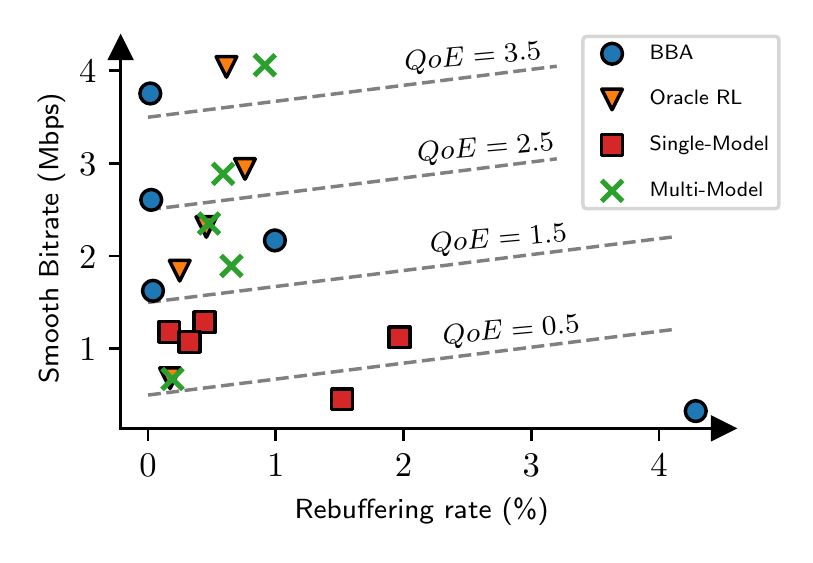}
    \caption{Break down of \gls{qoe} in \Cref{fig:abr_multi}; smooth bitrate refers to the chosen bitrate subtracted by the smoothness penalty.}
    \label{fig:app_abr_multi_scatter}
\end{figure}

\subsection{Extended results: QoE breakdown}

\label{app:abr_qoe_breakdown}

Recall that \gls{qoe} is a combination of several conflicting objectives. Namely, the quality of streaming, the smoothness of that quality across chunks and rebuffering:

$$QoE_t = q_t - |q_{t}-q_{t-1}| - \mu \cdot max(0, \frac{S_t}{c_t}-b_t)$$

To interpret the \gls{qoe} results in \Cref{fig:abr_multi}, we visualize the smoothed bitrate of choices against rebuffering. Smoothed bitrate is the combination of the first two terms in the \gls{qoe}. \Cref{fig:app_abr_multi_scatter} demonstrates this breaks down. \gls{bba} avoids rebuffering when bandwidth is stable at the cost of lower quality, but in variable bandwidths it repeats the same mistake of underestimating the risk of rebuffering. Single expert policies are confused by the non-stationary nature of the problem and break down in all cases. Multiple experts manage to avoid the confusion and \gls{cf} by design and adapt to any bandwidth distribution.

\subsection{Training setup}
\label{app:abr_train}
Similar to the straggler mitigation experiments, our implementation of \gls{a2c} uses the Pytorch~\cite{pytorch} library. We use the \gls{abr} implementation in~\cite{mao2019park} for our experiments. \Cref{table:abr_train} is a comprehensive list of all hyperparameters used in training and the environment. For the video in the \gls{abr} streaming, we use "Envivio-Dash3" from the DASH-246 JavaScript reference client~\cite{dash}.

\begin{table*}
\small
\centering
\begin{tabular}{c l l }
\toprule
\textbf{Group} & \textbf{Hyperparameter} & \textbf{Value} \\
\cmidrule{2-3}

\multirow{14}{*}{Neural network} & Hidden layers & (64, 32) \\
\cmidrule{2-3}
& Hidden layer activation function & Relu \\
\cmidrule{2-3}
& \multirow{2}{*}{Output layer activation function} & \gls{a2c} actor: Softmax \\
\cmidrule{3-3}
& & \gls{a2c} critic: Identity mapping\\
\cmidrule{2-3}
& Optimizer & Adam \cite{kingma2017adam} \\
\cmidrule{2-3}
& Learning rate & 0.001 \\
\cmidrule{2-3}
& $\beta_1$ & 0.9 \\
\cmidrule{2-3}
& $\beta_2$ & 0.999 \\
\cmidrule{2-3}
& $\epsilon$ & $10^{-8}$ \\
\cmidrule{2-3}
& Weight decay & $10^{-4}$ \\
\midrule

\multirow{8}{*}{\gls{a2c} training} & Episode lengths & 490 \\
\cmidrule{2-3}
& Epochs to convergence ($T_c)$ & 3000 (1470000 samples) \\
\cmidrule{2-3}
& Random seeds & 3 \\
\cmidrule{2-3}
& $\gamma$ & 0.96 \\
\cmidrule{2-3}
 & Entropy schedule & 0.25 to 0 in 2000 epochs \\
\cmidrule{2-3}
& $\lambda$ (for \gls{gae}) & 0.95 \\
\midrule

\multirow{2}{*}{Environment} & Chunk length $T$ & 4 \\
\cmidrule{2-3}
& Number of actions (bitrates) & 6 \\
\cmidrule{2-3}
& Fake-Replay starting threshold & $\min(20, 99^\text{th}~\text{of first 5 episode buffers})$ \\
\cmidrule{2-3}
& Fake-Replay schedule & From start to 0 in 2000 epochs \\
\midrule

\multirow{2}{*}{\gls{bba}} & Reservoir & 5 seconds (as in~\cite{mao2017neural}) \\
\cmidrule{2-3}
& Cushion & 10 seconds (as in~\cite{mao2017neural}) \\
\midrule
\bottomrule
\end{tabular}
\caption{Training setup and hyperparameters for \gls{abr} experiments.}
\label{table:abr_train}
\end{table*}
\end{appendices}

\end{document}